\documentclass[11pt]{article}

\usepackage[final]{acl}

\usepackage{times}
\usepackage{latexsym}

\usepackage[T1]{fontenc}

\usepackage[utf8]{inputenc}

\usepackage{microtype}

\usepackage{inconsolata}

\usepackage{xcolor}
\usepackage{graphicx}
\usepackage{colortbl}

\usepackage{booktabs}
\usepackage{multirow}
\usepackage{longtable}
\usepackage{array}
\usepackage{tabularx}
\usepackage{arydshln}
\usepackage{siunitx}
\usepackage[flushleft]{threeparttable}

\usepackage{amsmath}
\usepackage{amssymb}
\usepackage{amsthm}
\usepackage{stmaryrd}
\usepackage{algorithm}
\usepackage{algpseudocode}

\usepackage{hyperref}
\usepackage{url}
\usepackage{cleveref}

\usepackage{ragged2e}
\usepackage{csquotes}
\usepackage{enumitem}

\usepackage{listings}

\usepackage{pdflscape}
\usepackage{placeins}

\usepackage{caption}
\usepackage{subcaption}

\usepackage[most,skins,breakable]{tcolorbox}
\usepackage{stackengine}
\usepackage{bbding}
\usepackage{fancyvrb}

\usepackage{booktabs, multirow, makecell, rotating, xcolor, colortbl}
\usepackage[flushleft]{threeparttable}

\definecolor{grpA}{HTML}{E8F4FF}     % paler blue for group banners
\definecolor{grpB}{HTML}{FFE8EB}     % paler pink for group banners
\definecolor{grpC}{HTML}{E8FFEF}     % paler mint for group banners
\definecolor{grpD}{HTML}{FFF0E0} 
\definecolor{grpE}{HTML}{FFF0F6} 
\definecolor{avgcol}{gray}{0.95}      % avg column grey
\definecolor{dscol}{gray}{0.95}      % dataset name column background
\definecolor{latcol}{RGB}{210,235,255} % light blue for latency column
\definecolor{latcpucol}{RGB}{210,235,255} % light blue (CPU)
\definecolor{latgpucol}{RGB}{170,215,255} % slightly darker blue (GPU)

\usepackage{pifont}
\usepackage{wrapfig}

\title{ViDoRe V3: A Comprehensive Evaluation of Retrieval Augmented Generation in Complex Real-World Scenarios} % problème d'anonymisation 

\newcommand{\authorlink}[2]{\href{mailto:#1}{\textcolor{black}{\textbf{#2}}}}

\author{
\authorlink{antonio.loison@illuin.tech}{António Loison\thanks{Equal contribution}\thanks{Work done while at Illuin Technology}}
    \quad \authorlink{quentin.mace@illuin.tech}{Quentin Macé$^*$$^{1,3}$}
    \quad \authorlink{antoine.edy@illuin.tech}{Antoine Edy$^*$$^{1}$} \\
    \authorlink{victor.xing@illuin.tech}{Victor Xing$^{1}$}
    \quad \authorlink{tbalough@nvidia.com}{Tom Balough$^{2}$}
    \quad \authorlink{gmoreira@nvidia.com}{Gabriel Moreira$^{2}$} 
    \quad \authorlink{boli@nvidia.com}{Bo Liu$^{2}$}  \\
    \authorlink{manuel.faysse@centralesupelec.fr}{Manuel Faysse$^{3}$\textsuperscript{\dag}}
    \quad \authorlink{celine.hudelot@centralesupelec.fr}{Céline Hudelot$^{3}$}
    \quad \authorlink{gautier.viaud@illuin.tech}{Gautier Viaud$^{1}$} \\
     $^{1}$Illuin Technology
     \quad$^2$NVIDIA
     \quad$^3$CentraleSupélec, Paris-Saclay \\
     \small\texttt{\{quentin.mace, antoine.edy, gautier.viaud\}@illuin.tech\thanks{Contact emails}}
    }

\definecolor{superlightgray}{gray}{0.95}
\definecolor{borderblack}{gray}{0.1}

\newenvironment{promptfigure}[3][superlightgray]{%
    \begin{figure*}[h!]
    \centering
    \def\promptCaption{#2}%
    \def\promptLabel{#3}%
    \begin{tcolorbox}[
        width=\textwidth,
        colback=#1,
        colframe=borderblack,
        boxrule=0.5pt,
        arc=3mm,
        left=4mm, right=4mm, top=4mm, bottom=4mm,
        fonttitle=\bfseries\Large,
        fontupper=\small,
        parskip=0.5\baselineskip
    ]
}{%
    \end{tcolorbox}
    \caption{\promptCaption}
    \label{\promptLabel}
    \end{figure*}
}

\definecolor{ownpurple}{RGB}{111, 8, 201}
\definecolor{sportingreen}{RGB}{0, 128, 87}
\definecolor{ownblue}{RGB}{68, 140, 255}
\definecolor{borderblack}{rgb}{0.1, 0.1, 0.1} % Darker black for the frame
\definecolor{superlightgray}{rgb}{0.97, 0.97, 0.97} % Very light background

\newcommand{\vidorev}{ViDoRe~V3}

\def\vidore{ViDoRe}
\def\vidorev{\mbox{ViDoRe V3}}

\begin{document}
\maketitle
\begin{abstract}
Retrieval-Augmented Generation (RAG) pipelines must address challenges beyond simple single-document retrieval, such as interpreting visual elements (tables, charts, images), synthesizing information across documents, and providing accurate source grounding. Existing benchmarks fail to capture this complexity, often focusing on textual data, single-document comprehension, or evaluating retrieval and generation in isolation. We introduce \textbf{\vidorev}, a comprehensive multimodal RAG benchmark featuring multi-type queries over visually rich document corpora. It covers 10 datasets across diverse professional domains, comprising ~26,000 document pages paired with 3,099 human-verified queries, each available in 6 languages. Through 12,000 hours of human annotation effort, we provide high-quality annotations for retrieval relevance, bounding box localization, and verified reference answers. Our evaluation of state-of-the-art RAG pipelines reveals that visual retrievers outperform textual ones, late-interaction models and textual reranking substantially improve performance, and hybrid or purely visual contexts enhance answer generation quality. However, current models still struggle with non-textual elements, open-ended queries, and fine-grained visual grounding. To encourage progress in addressing these challenges, the benchmark is released under a commercially permissive license\footnote{\url{https://hf.co/vidore}}. % and is linked in the public version of this work.

\end{abstract}

\section{Introduction}

\begin{figure}[t!]
    \centering
    \includegraphics[trim={3cm 7.5cm 3cm 4cm}, width=0.73\columnwidth]{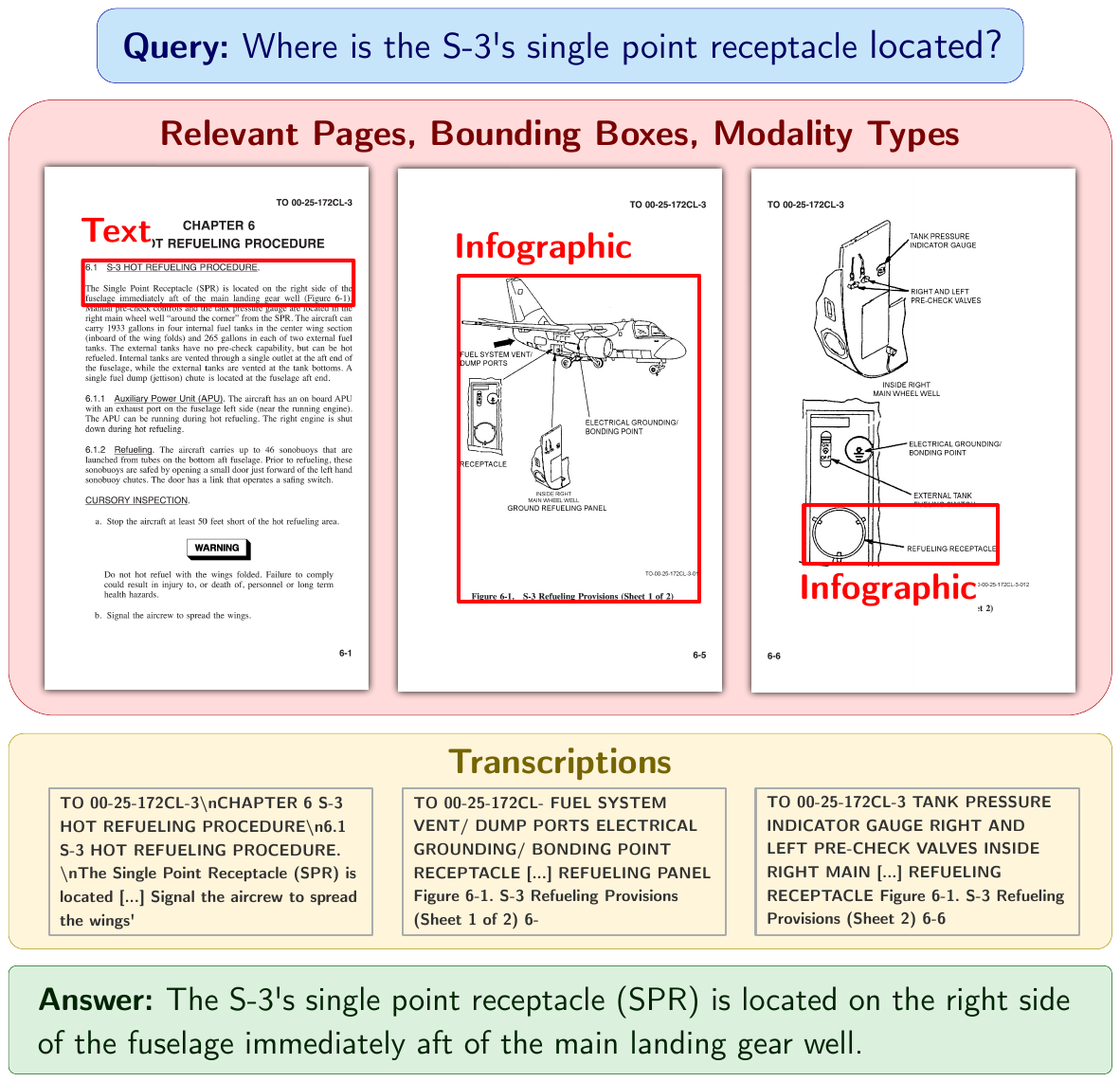}
    \caption{\textbf{ViDoRe V3 sample.} Each query is annotated with the relevant pages, a document-grounded answer, bounding boxes localizing supporting evidence and modality labels for each bounding box. Documents are provided in image, text and PDF formats.} % TODO: clean this caption
    \label{fig:sample}
\end{figure}

Retrieval-Augmented Generation (RAG)~\citep{lewis2021retrievalaugmentedgenerationknowledgeintensivenlp} has become the dominant paradigm for knowledge-intensive NLP tasks~\citep{gao2024retrievalaugmentedgenerationlargelanguage, fan2024surveyragmeetingllms}. Yet practical deployments introduce complexities that academic benchmarks often overlook when focusing on single-document textual retrieval. First, documents encode critical information in visual elements such as tables, charts, and images designed for human interpretation, which text-only pipelines often ignore~\citep{abootorabi2025askmodalitycomprehensivesurvey, cho2024m3docragmultimodalretrievalneed}. Second, user queries often require open-ended synthesis, comparison, and reasoning over scattered information, not simple factoid lookup~\citep{tang2024multihopragbenchmarkingretrievalaugmentedgeneration, conti2025contextgoldgoldpassage, thakur2025freshstackbuildingrealisticbenchmarks}. Third, trustworthy systems must ground responses to specific source locations (e.g., bounding boxes), to mitigate hallucinations~\citep{gao2023enablinglargelanguagemodels, ma2024visaretrievalaugmentedgeneration}.

Existing benchmarks leave these requirements only partially addressed. Early Visual Document Understanding (VDU) benchmarks focus on single-page comprehension, ignoring the complexity of large document corpora~\citep{mathew2021docvqa}. Recent retrieval-centric benchmarks do not evaluate generation quality and grounding~\citep{faysse2025colpaliefficientdocumentretrieval,günther2025jinaembeddingsv4universalembeddingsmultimodal}. Some multimodal datasets attempt to bridge this gap but rely on extractive, short-answer tasks that fail to exercise complex reasoning~\citep{cho2024m3docrag}, or lack multilingual diversity and fine-grained visual grounding~\citep{peng2025unidocbenchunifiedbenchmarkdocumentcentric}.

\begin{figure*}[t!]
    \centering
    \includegraphics[width=\textwidth]{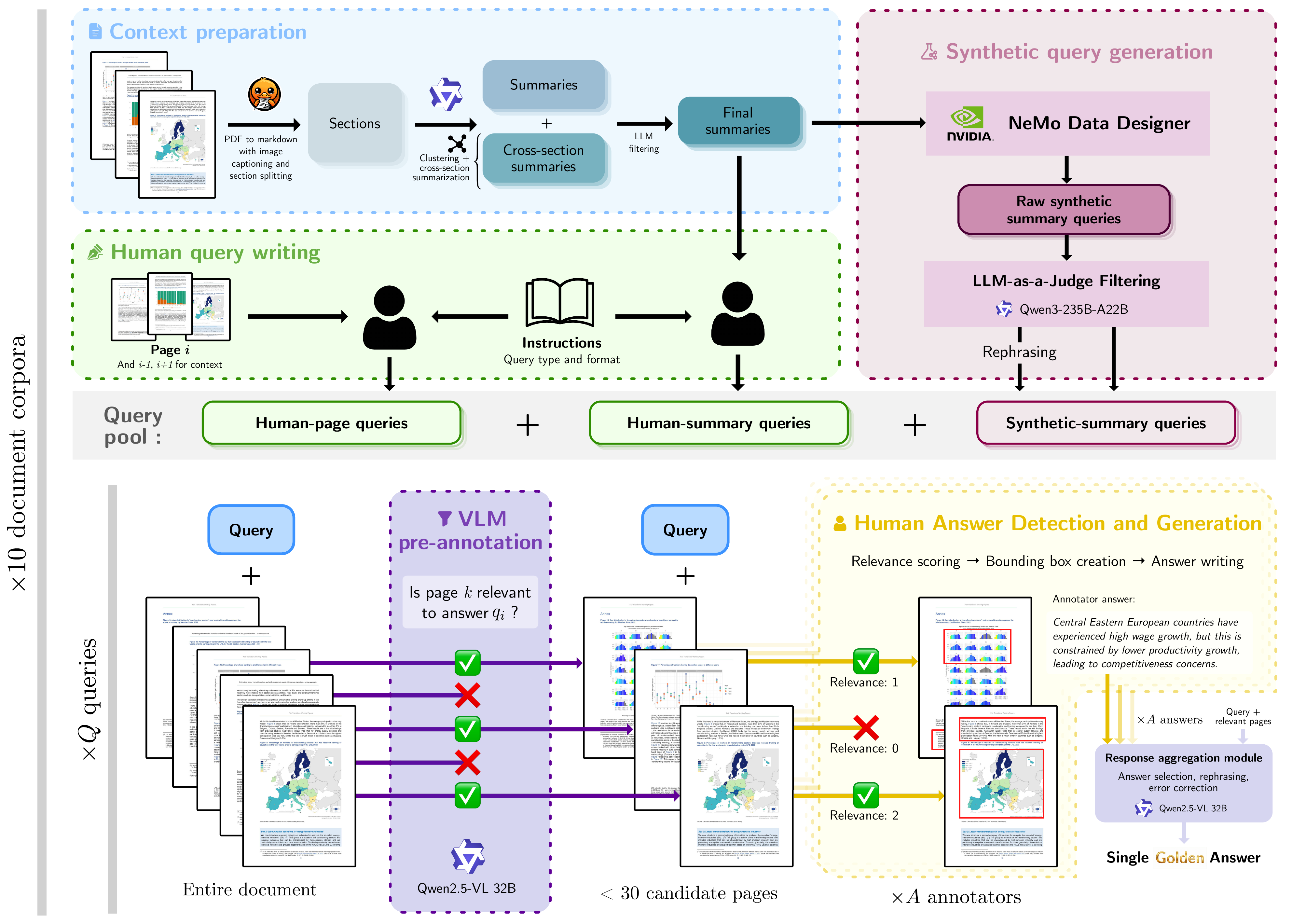}
    \caption{\textbf{Overview of the benchmark creation process.} Queries are sourced from 3 streams: \textit{human extractive} (using raw pages), \textit{human blind contextual} (using summaries to mitigate extractive bias), and \textit{synthetic blind contextual}. For each query, a VLM pre-filtered subset of candidate pages is labeled by 1--3 human annotators that perform relevance scoring, bounding box localization and answer generation. A final response aggregation combines annotator answers into a single answer.} %\gmor{Could you identify in the main figure what are Tasks 1 and Task 2?}\anto{Yes, it would be great to reuse the terms Query Writing and Query Linking}}
    \label{fig:full process}
\end{figure*}

To address these limitations, we introduce \textbf{\vidorev}, a benchmark designed for complex and realistic end-to-end RAG evaluation on visually rich document corpora. Our contributions are:

\paragraph{1. A Human Annotation Methodology for Realistic Queries} We propose an annotation protocol for generating diverse queries and fine-grained query-page annotations. By restricting annotator access to document content during query formulation, we capture authentic search behaviors and mitigate bias toward simple extractive queries. Vision-Language Model (VLM) filtering combined with human expert verification enables efficient, high-quality annotation at scale.

\paragraph{2. The ViDoRe V3 Benchmark} Applying this methodology to 10 industry-relevant document corpora, we build ViDoRe V3, a multilingual RAG benchmark comprising ~26,000 pages and 3,099 queries, each available in 6 languages. Two datasets are held out as a private test set to mitigate overfitting. The benchmark is fully integrated into the MTEB ecosystem and \mbox{leaderboard}\footnote{\href{https://mteb-leaderboard.hf.space/?benchmark_name=ViDoRe\%28v3\%29}{https://mteb-leaderboard.hf.space}}~\citep{muennighoff2023mteb}, and the public datasets are released under a commercially permissive license.  %The benchmark is fully integrated into the MTEB ecosystem and leaderboard~\citep{muennighoff2023mteb}, and the public datasets are released under a commercially permissive license.
\paragraph{3. Comprehensive Evaluation and Insights} Leveraging our granular annotations, we benchmark state-of-the-art models on \textbf{(i)} retrieval accuracy by modality and language, \textbf{(ii)} answer quality across diverse retrieval pipeline configurations, and \textbf{(iii)} visual grounding fidelity. Our analysis surfaces actionable findings for RAG practitioners.

\section{Related Work}

\paragraph{Component-Level Benchmarks (VDU and Retrieval)} VDU has traditionally relied on single-page datasets like DocVQA~\citep{mathew2021docvqa}, alongside domain-specialized variants~\citep{mathew2021infographicvqa, zhu2022towards, wang2024charxiv}. These ignore the multi-page context inherent to RAG. Recent work evaluating bounding-box source grounding~\citep{yu2025bboxdocvqalargescale} proposes single-page and multi-page tasks but does not address the retrieval component. Conversely, the emergence of late-interaction visual retrievers~\citep{ma2024unifyingmultimodalretrievaldocument, faysse2025colpaliefficientdocumentretrieval, yu2025visragvisionbasedretrievalaugmentedgeneration, xu2025llamanemoretrievercolembedtopperforming} spurred the creation of retrieval-centric visual benchmarks like Jina-VDR ~\citep{günther2025jinaembeddingsv4universalembeddingsmultimodal} and \vidore \space V1\&V2 ~\citep{ faysse2025colpaliefficientdocumentretrieval, macé2025vidorebenchmarkv2raising}, but none of these benchmarks jointly evaluate retrieval and answer generation.

\paragraph{End-to-End Multimodal RAG} While recent textual RAG benchmarks now capture complex user needs like reasoning or summarizing~\citep{thakur2025freshstackbuildingrealisticbenchmarks, tang2024multihopragbenchmarkingretrievalaugmentedgeneration, su2024bright}, multimodal evaluation often remains limited to single page queries~\citep{faysse2025colpaliefficientdocumentretrieval}. Multi-page datasets like DUDE~\citep{van2023document}, M3DocRAG~\citep{cho2024m3docragmultimodalretrievalneed}, ViDoSeek~\citep{wang2025vidorag} or Real-MM-RAG \citep{wasserman2025real} prioritize extractive retrieval, lacking the diversity of queries encountered in realistic settings. 
UniDocBench~\citep{peng2025unidocbenchunifiedbenchmarkdocumentcentric} represents a concurrent effort that similarly addresses diverse query types and provides comparative evaluation across multiple RAG paradigms. While this benchmark offers valuable contributions, it relies on synthetically generated queries via knowledge-graph traversal, is restricted to English documents, and constrains grounding annotations to parsed document elements. In contrast, our benchmark offers several complementary strengths: fully human-verified annotations, a cross-lingual setup, free-form bounding box annotations, and a more systematic evaluation of individual visual RAG pipeline components.

\section{Benchmark Creation} \label{sec:benchmark creation}

We design the benchmark to mirror the diversity of information retrieval situations in large-scale realistic environments. To enable pipeline-agnostic evaluation of the 3 core RAG components (retrieval, generation and grounding), while avoiding limitations of synthetic benchmarks, we employ a rigorous three-stage human-in-the-loop annotation process involving document collection, query generation and grounded query answering (\Cref{fig:full process}).

\subsection{Document Collection}

We curate 10 diverse corpora by manually selecting openly-licensed documents from governmental, educational, and industry sources, focusing on English and French documents (7 and 3 corpora respectively). The corpora span Finance, Computer Science, Energy, Pharmaceuticals, Human Resources, Industrial Maintenance, Telecom, and Physics. Each features domain-specific terminology and document structures representative of realistic retrieval tasks (details in \Cref{tab:datasets}).

\subsection{Query Generation}

\paragraph{Query Taxonomy} To evaluate document visual retrieval systems across diverse realistic scenarios, we develop a query taxonomy with two orthogonal dimensions: \textit{Query Type}, defining the user's information need, and \textit{Query Format}, describing the query's syntactic structure. This dual-axis classification enables more nuanced performance analysis than benchmarks focusing solely on interrogative extractive queries. We define 7 Query Types: \textit{open-ended}, \textit{extractive}, \textit{numerical}, \textit{multi-hop}, \textit{compare-contrast}, \textit{boolean}, and \textit{enumerative}, and 3 Query Formats: \textit{question}, \textit{keyword}, and \textit{instruction}. % that reflect modern search engine usage \vic{last part to justify or remove}\footnote{Taxonomy details are described in \Cref{tab:query_taxonomy}}. % To generate even more diverse queries, a synthetic and human query generation processes were followed.

\paragraph{Context Preparation} We further ensure query diversity by pulling summaries from a heterogeneous set of contexts during the generation process.
Two types of input contexts are used: specific document sections that target local information retrieval and cross-section summaries that target multi-document context retrieval. These summaries are produced through a refined process inspired by ViDoRe V2~\cite{macé2025vidorebenchmarkv2raising}. First, the text is extracted from PDFs using Docling~\cite{auer2024docling} along with image descriptions. Then, summaries are generated with Qwen3-235B-Instruct~\cite{qwen3technicalreport} from each document section. They are clustered to group similar summaries together using Qwen3-Embedding-0.6B \cite{qwen3embedding} as embedder, UMAP \cite{mcinnes2020umapuniformmanifoldapproximation} for dimension reduction and HDBSCAN \cite{campello2013density} for clustering. Additionally, cross-section summaries are produced by synthesizing the summaries of 2 to 3 randomly selected sections per cluster. From this pool of summaries, a final subset is curated to maintain a strict balance between single-section and cross-section summaries. The selection also ensures an even distribution across section modalities (text, images, and tables) as defined by the Docling element classification.

\paragraph{Synthetic Query Generation} Queries are generated from the summaries using a first synthetic generation pipeline based on Qwen3-235B. For each summary, a prompt is constructed by sampling a query type and format at random, together with variable attributes such as length and difficulty, in order to promote diversity. The generated queries are subsequently evaluated by the same LLM acting as an automatic judge, which filters outputs according to 4 criteria: information richness, domain relevance, clarity and adherence to query type/format. Finally, 50\% of the retained queries are rephrased to further enhance linguistic variance. This pipeline is implemented using NeMo Data Designer \cite{nemo-data-designer} to facilitate generation scaling.

\paragraph{Human Query Writing} Human annotators are provided 2 kinds of contexts: synthetic summaries or specific PDF pages. They are tasked with generating one query following a specific query type and format and one query of their choice that is most adapted to the context provided.

\subsection{Answer Detection and Generation} \label{sec: grounded qa}

Queries are filtered and linked to relevant pages using a hybrid pipeline of VLM pre-filtering and human annotation. It is followed by human answer annotation and visual grounding.

\paragraph{Query-Page Linking} Given the scale of our corpora, manual verification of each page relevance for each query is intractable. We therefore adopt a two-stage annotation pipeline. First, Qwen2.5-VL-32B-Instruct \citep{qwen25vl} pre-filters candidate pages by assessing whether each page image is relevant to the query. Queries whose answers span more than 30 flagged pages are discarded. Human annotators then review the remaining query-page pairs, evaluating query quality and rating page relevance on a three-point scale (Not Relevant, Critically Relevant, Fully Relevant). We selected Qwen2.5-VL-32B-Instruct for its high recall, prioritizing coverage over precision and leaving final relevance judgments entirely to human annotators (see Appendix~\ref{appendix: vlm filtering stats} for details and distributional validation).

\paragraph{Relevant Page Selection} To ensure annotation quality, each task is completed by multiple annotators and reviewed by annotation supervisors. Since VLM pre-filtering biases the distribution toward relevant pages, we report Gwet's AC2, as it remains stable under prevalence skew, at 0.760 (see \Cref{appendix: relevance agreement} for dataset-level breakdowns). Given this strong but imperfect agreement, we implement a tiered review process: extractive queries require at least one annotator and one reviewer, while more complex non-extractive queries require at least two annotators and one reviewer. A page is retained as relevant if marked by either (i) one annotator and one reviewer, or (ii) at least two annotators.  %\manu{to me this is a detail that should go in app. here we can say double annotation is used to guarantee quality.}\anto{probably to remove}

\paragraph{Answer Generation} For each selected query, annotators were tasked with writing an answer based on the pages they marked as relevant. Given that different annotators might have different answer interpretations and tend not to be exhaustive in their answers, we use Qwen2.5-VL-32B-Instruct to generate a final answer based on the relevant page images marked by the annotators and their answers. To validate that this aggregation faithfully preserves annotator intent, we evaluated the VLM-aggregated answer against individual annotator responses using a GPT-5.2 judge to assess factual consistency. The aggregated answer matched the exact informational content (with minor paraphrasing) of at least one annotator's response in 86.3\% of cases, confirming the VLM predominantly acts as a selector. For the remaining 13.7\% of divergent cases, manual review of a random subset showed the aggregated answer was judged superior in most cases, either by merging complementary information from two incomplete responses or by correcting verifiable factual errors in individual annotations.

\paragraph{Bounding Boxes and Modality Types}
For each relevant page, annotators delineate bounding boxes around content supporting the query and attribute a modality type to each bounding box: Text, Table, Chart, Infographic, Image, Mixed or Other. 
Because multiple valid interpretations of bounding boxes can exist, we perform a consistency study to evaluate inter-annotator agreement and establish a human performance upper bound for the task.

We compute inter-annotator agreement on the subset of query-page pairs labeled by two or three annotators. For each annotator, we merge all their bounding boxes into a single zone. We then compare zones across annotators by measuring pixel-level overlap, reporting Intersection over Union (IoU) and F1 score (Dice coefficient). When three annotators label the same sample, we average over all pairwise comparisons.

Across all 10 datasets, we observe an average IoU of 0.50 and F1 of 0.60. These moderate agreement scores reflect the inherent subjectivity of the task: annotators typically agreed on the relevant content but differed in granularity (Appendix \ref{appendix: bbox examples}), with some marking tight bounds around specific content while others included surrounding context. Section~\ref{sec:visual grounding results} describes how our evaluation methodology accounts for this ambiguity and how model scores should be interpreted relative to this human ceiling.

\paragraph{Quality Control} The annotation was conducted by a curated pool of 76 domain-qualified experts with native-level language proficiency. Quality control was performed by 13 senior annotators with enhanced domain knowledge and extensive annotation experience. Detailed protocols regarding the annotator pool and training are provided in Appendix \ref{appendix: annotator pool}.

\subsection{Final Query Distribution} 

\begin{figure}[ht]
    \centering
    \includegraphics[width=\columnwidth]{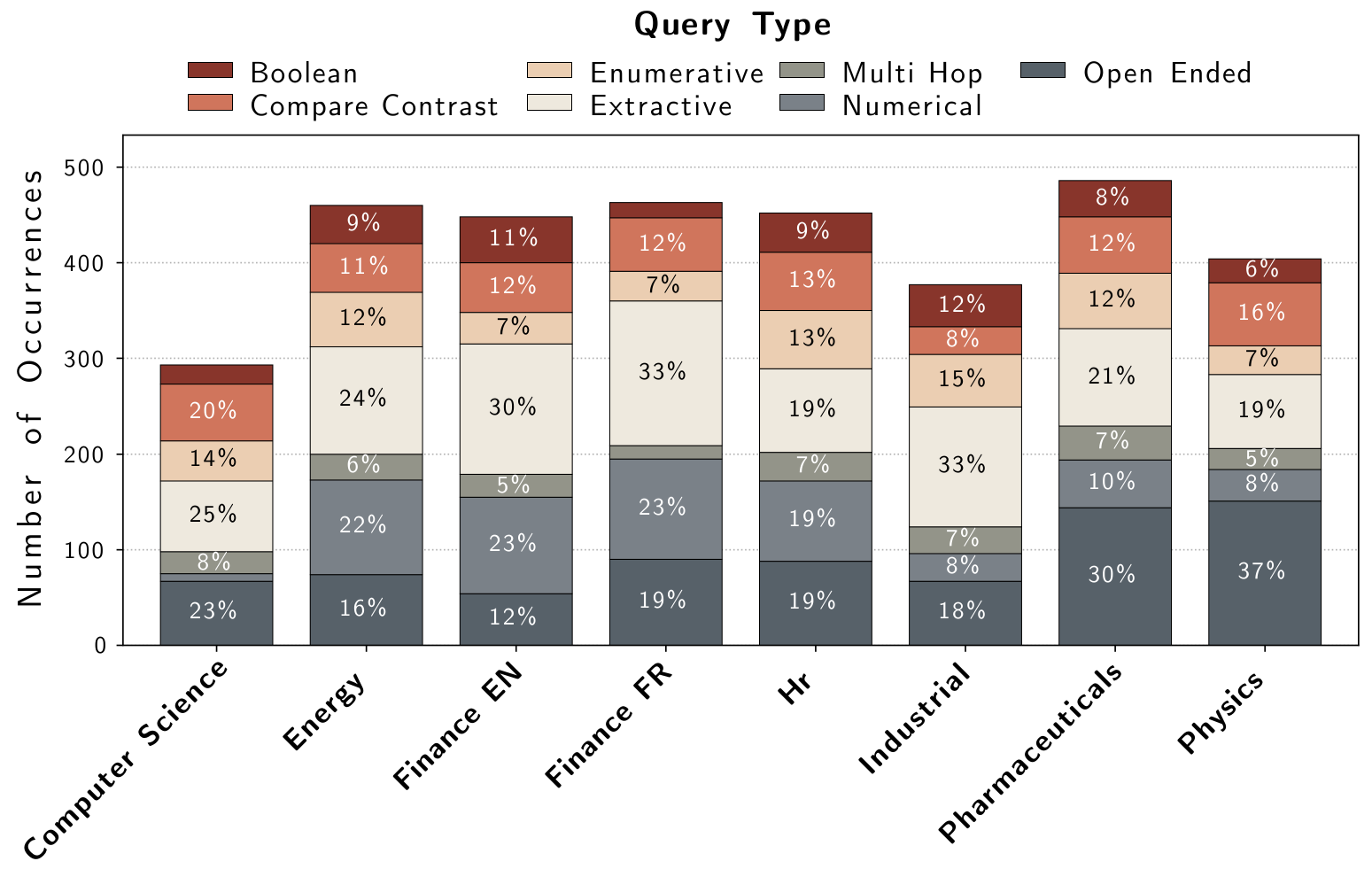}
    \caption{\textbf{Query Type Distribution per Domain}} 
    \label{fig:query_type_distribution}
\end{figure}

\begin{figure}[ht]
    \centering
    \includegraphics[width=\columnwidth]{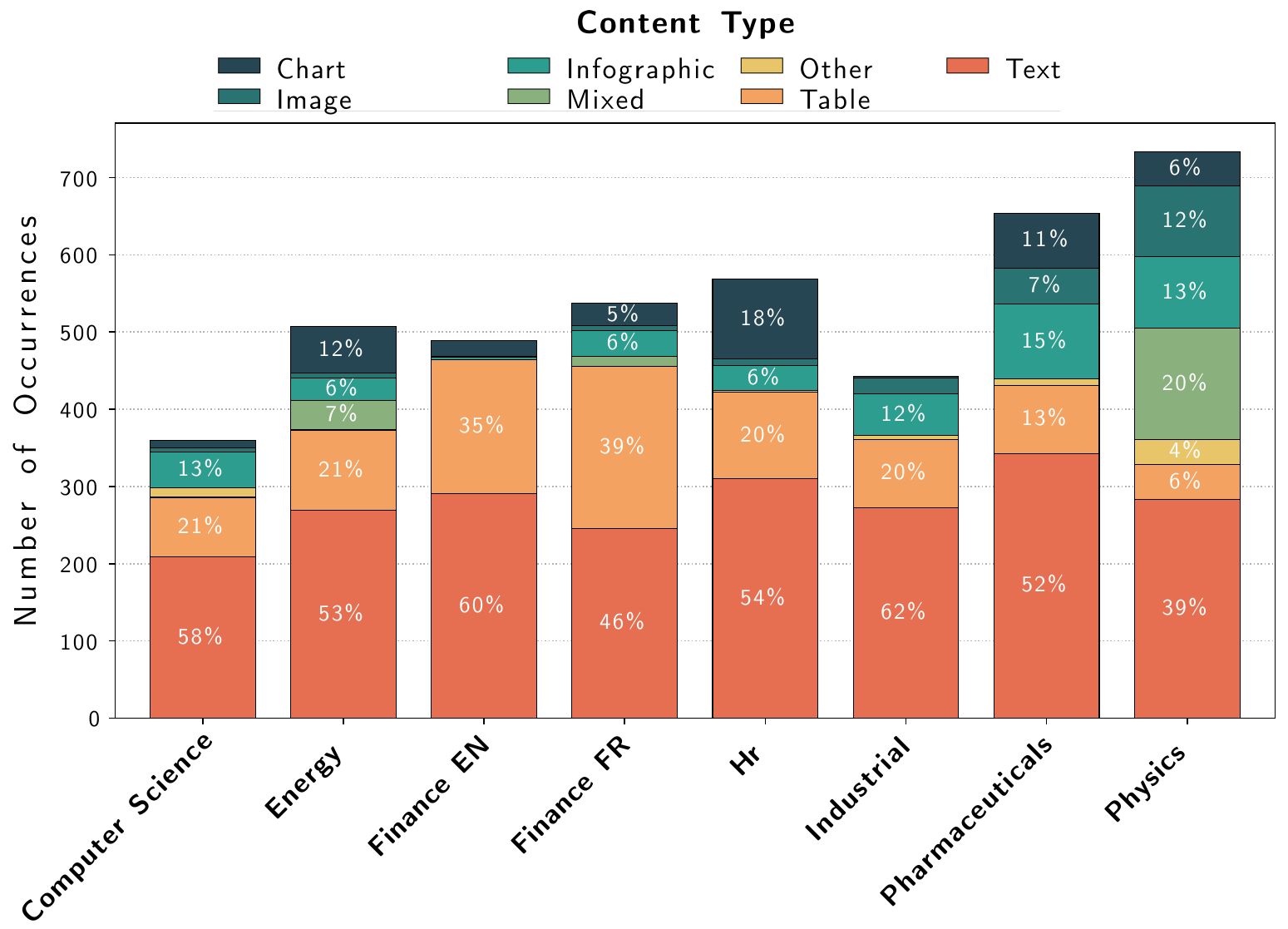}
    \caption{\textbf{Content Type Distribution per Domain}} 
    \label{fig:content_type_distribution}
\end{figure}

We conducted a final human review to remove low-quality queries and resolve labeling ambiguities. \Cref{fig:query_type_distribution} shows the resulting distribution. Extractive queries predominate due to human annotator preference, followed by open-ended queries from targeted sampling. Multi-hop queries were the hardest to scale, suggesting a need for dedicated pipelines. \Cref{fig:content_type_distribution} details page modalities; while text is most prevalent, visual elements like tables, charts, and infographics are well-represented.

\subsection{Dataset Release and Distribution}

We extend the benchmark to rigorously assess cross-lingual retrieval. While source documents are maintained in English and French, we use Qwen3-235B-Instruct to provide translations in 6 languages: English, French, Spanish, German, Italian, and Portuguese. This configuration challenges models to bridge the semantic gap between the query language and the document language, a critical requirement for modern RAG systems. 

Finally, to ensure the integrity of evaluation and mitigate data contamination (which was shown to be a major preoccupation for Information Retrieval \cite{rteb2025}), we adopt a split-release strategy. 8 datasets are made public to facilitate research, while 2 are retained as private hold-out sets. This enables blind evaluation, ensuring that performance metrics reflect true generalization rather than overfitting to public samples.

\section{Experiments and Results}

Using our benchmark, we conduct extensive evaluations across all 3 components of RAG pipelines. We assess textual and visual retrievers and rerankers on retrieval performance, evaluate leading VLMs and LLMs on their ability to generate accurate answers from various retrieved contexts, and test VLMs on bounding box generation for visual grounding. From these results, we compile practical insights for RAG practitioners.

\begin{table*}[h!]
\small
\centering
\resizebox{\textwidth}{!}{%
\begin{tabular}{l c c c c c c c c c c c >{\columncolor{avgcol}}c}
\toprule
 & & \multicolumn{7}{c}{\textbf{English Datasets}} & \multicolumn{3}{c}{\textbf{French Datasets}} & \multicolumn{1}{c}{} \\
\cmidrule(lr){3-9} \cmidrule(lr){10-12}
\textbf{Model} & Size (B) & \shortstack{C.S.} & \shortstack{Nucl.} & \shortstack{Fin.} & \shortstack{Phar.} & \shortstack{H.R.} & \shortstack{Ind.} & \shortstack{Tel.} & \shortstack{Phys.} & \shortstack{Ener.} & \shortstack{Fin.} & \textbf{Avg.} \\
\midrule
\multicolumn{13}{l}{\textit{\textbf{Textual Retrievers}}} \\
Qwen3-8B$^\bigstar$  & 8 & \textbf{71.7} & 39.0 & \textbf{49.4} & \textbf{59.2} & 47.6 & \textbf{40.4} & \textbf{62.8} & \textbf{45.6} & 58.9 & 35.8 & \textbf{51.0} \\
Jina-v4 & 3 & 64.3 & \textbf{44.3} & 48.4 & 54.9 & \textbf{52.8} & 38.4 & 56.3 & 43.6 & \textbf{60.1} & \textbf{41.3} & 50.4 \\
LFM2-350M & 0.35 & 63.5 & 37.8 & 39.0 & 56.4 & 43.5 & 34.4 & 56.9 & 41.8 & 47.0 & 28.2 & 44.9 \\
Qwen3-0.6B$^\bigstar$  & 0.6 & 66.4 & 32.8 & 42.7 & 50.6 & 37.7 & 31.6 & 55.7 & 43.3 & 51.3 & 25.8 & 43.8 \\
BGE-M3$^\bigstar$  & 0.57 & 58.0 & 30.2 & 39.8 & 52.0 & 42.4 & 28.5 & 51.6 & 35.9 & 49.8 & 25.2 & 41.3 \\
BM25S & - & 28.7 & 17.4 & 17.6 & 27.3 & 12.8 & 15.6 & 33.3 & 14.8 & 21.9 & 14.0 & 20.3 \\
\midrule
\multicolumn{13}{l}{\textit{\textbf{Visual Retrievers}}} \\
ColEmbed-3B-v2 & 3 & \textbf{77.1} & \textbf{50.7} & \textbf{64.2 }& \textbf{66.0} & \textbf{62.3} & \textbf{51.7} & \textbf{69.7} & 47.0 & 64.9 & 44.4 & \textbf{59.8} \\
Jina-v4 & 3 & 71.8 & 50.0 & 59.3 & 63.1 & 59.5 & 50.4 & 64.8 & 46.6 & 64.0 & \textbf{46.1} & 57.6 \\
ColNomic-7B & 7 & 76.2 & 45.0 & 56.6 & 62.3 & 58.7 & 50.1 & 67.2 & \textbf{48.3} & 64.0 & 45.5 & 57.4 \\
ColEmbed-3B & 3 & 75.2 & 49.1 & 60.9 & 63.7 & 58.7 & 47.1 & 67.0 & 45.1 & 62.1 & 43.8 & 57.3 \\
ColNomic-3B & 3 & 72.7 & 42.1 & 56.3 & 61.1 & 57.3 & 47.4 & 64.5 & 47.5 & \textbf{65.0} & 44.3 & 55.8 \\
ColEmbed-1B & 1 & 71.3 & 47.3 & 58.9 & 62.6 & 57.0 & 46.6 & 64.7 & 44.1 & 60.9 & 42.4 & 55.6 \\
ColQwen2.5 & 3 & 72.3 & 38.1 & 52.3 & 57.9 & 51.2 & 41.3 & 61.3 & 45.9 & 59.7 & 39.1 & 51.9 \\
Nomic-7B$^\bigstar$  & 7 & 66.6 & 36.7 & 48.8 & 58.9 & 46.2 & 37.9 & 57.8 & 44.2 & 57.5 & 36.0 & 49.0 \\
ColQwen2 & 2 & 68.6 & 35.7 & 39.0 & 52.2 & 45.1 & 38.3 & 57.4 & 41.6 & 48.8 & 20.0 & 44.7 \\
Nomic-3B$^\bigstar$ & 3 & 58.5 & 32.2 & 44.2 & 55.3 & 43.3 & 33.2 & 53.7 & 42.0 & 51.4 & 28.9 & 44.3 \\
ColPali & 7 & 65.3 & 32.9 & 34.4 & 53.1 & 44.8 & 35.6 & 54.0 & 41.7 & 47.1 & 21.8 & 43.1 \\
\bottomrule
\end{tabular}
}
\caption{\textbf{Retrieval performance (NDCG@10) across the benchmark.} Best results per category in bold. \mbox{$\bigstar$: single-vector models}. Following MTEB conventions, the average score is a macro-average over all datasets. Full model names and references are found in Table~\ref{tab:model_metadata}.} %\boliu{FYI: I updated the full name of ColEmbed-3B-v2 in Table 9, to be consistent to ColEmbed-3B-v1} \anto{thank you}
\label{tab:retriever_comparison_updated}
\end{table*}

\subsection{Retrieval} \label{sec:retrieval results}

We evaluate a large panel of visual and textual retrievers on page-level retrieval ability. Visual retrievers are given page images, while textual retrievers process the Markdown text of each page processed by the NeMo Retriever extraction service\footnote{Chunking within pages or providing image descriptions did not improve our results. Thus, we report the results of the simplest pipeline.}~\cite{nvidiaocr}. The results reported in Table~\ref{tab:retriever_comparison_updated} corroborate findings from existing document retrieval benchmarks~\cite{faysse2025colpaliefficientdocumentretrieval,günther2025jinaembeddingsv4universalembeddingsmultimodal}: for a given parameter count, visual retrievers outperform textual retrievers, and late interaction methods score higher than dense methods.

We analyze ColEmbed-3B-v2, the best-performing retriever we evaluated across query type, content modality, and query language. % detailed results are reported in Appendix~\ref{appendix:supp retrieval}.
A breakdown by query generation source is provided in Appendix~\ref{appendix:supp retrieval} (Table~\ref{tab:perf_query_source}).

\begin{figure}[ht]
    \centering
    \includegraphics[width=0.99\columnwidth]{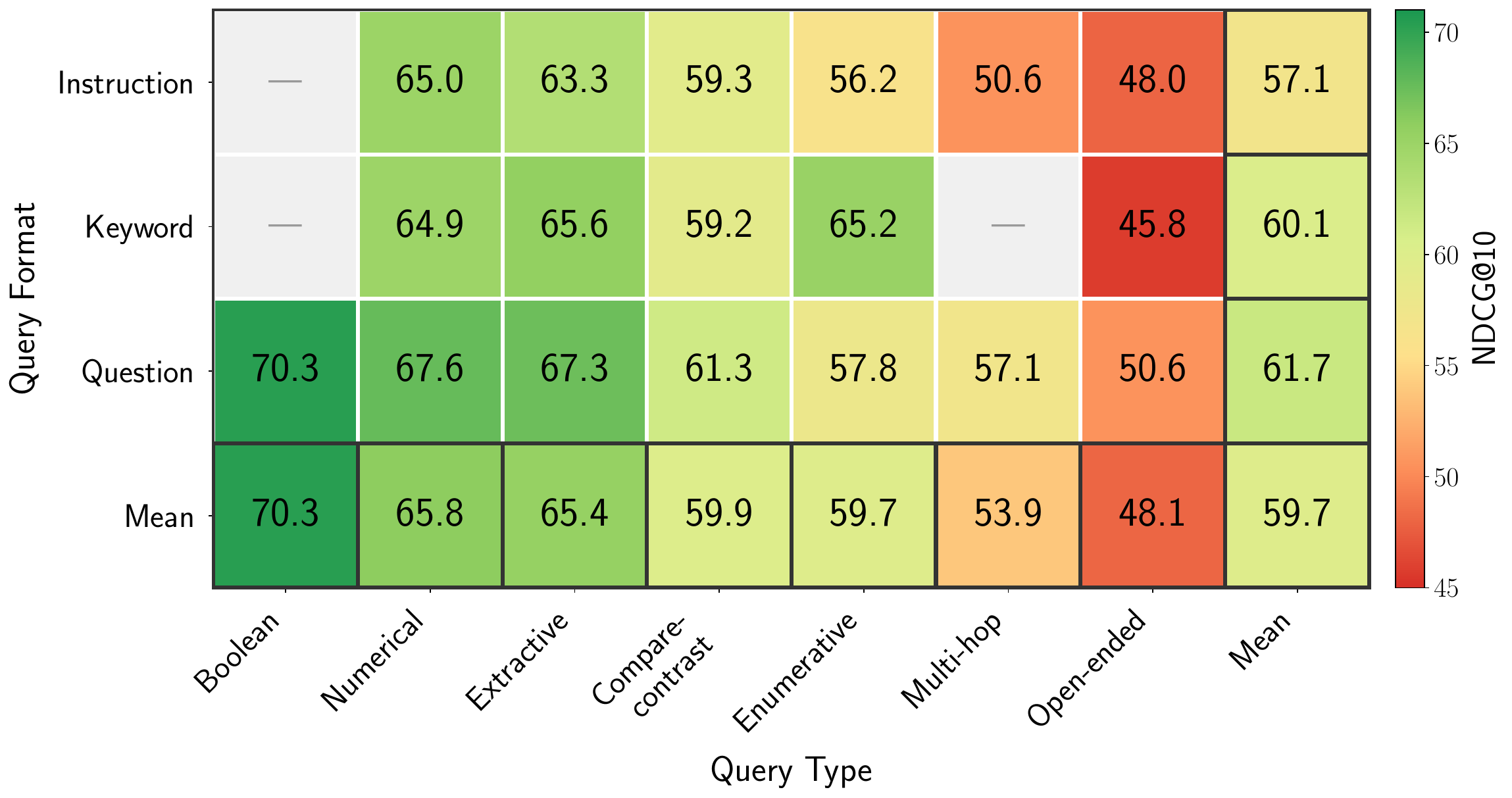}
    \caption{\textbf{ColEmbed-3B-v2 NDCG@10 by query type and format.}}
    \label{fig:type-format-performance}
\end{figure}

\paragraph{Performance is aligned with query complexity}
\Cref{fig:type-format-performance} shows that performance is inversely correlated with query complexity: simple query types such as Boolean and Numerical score significantly higher than Open-ended and Multi-hop queries. Question formulations consistently outperform Instruction and Keyword formats across nearly all categories, underscoring the need for improved handling of these query structures. %, with the notable exception of the Enumerative type for which Keyword queries score 7.4 points higher.

\begin{figure}[ht]
    \centering
    \includegraphics[width=\columnwidth]{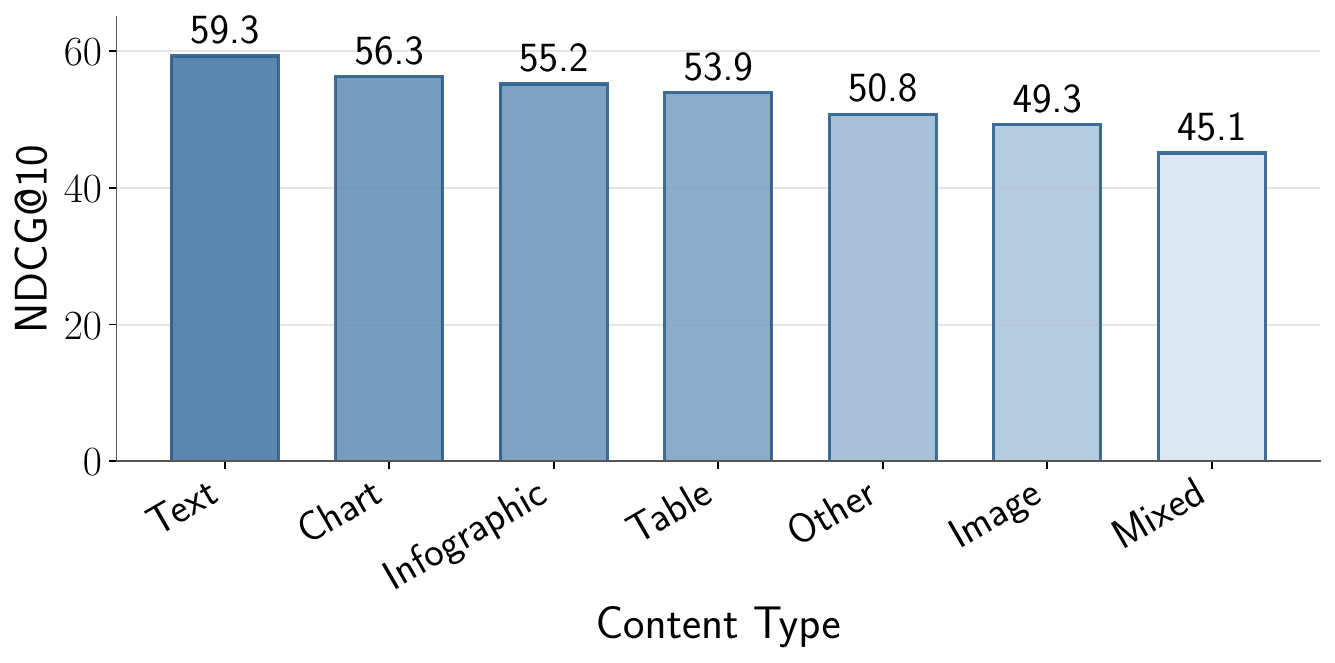}
    \caption{\textbf{ColEmbed-3B-v2 NDCG@10 by modality.}} 
    \label{fig:content-type-performance}
\end{figure}

\begin{figure}[h!]
    \centering
    \includegraphics[width=\columnwidth]{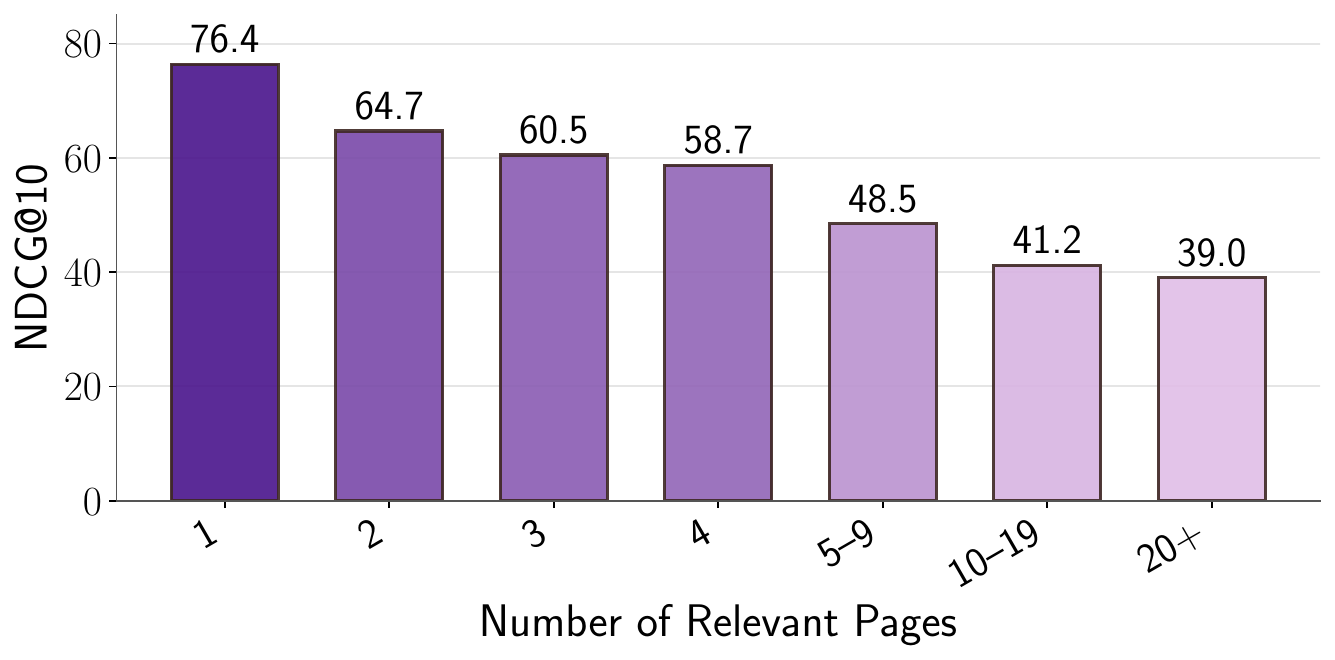}
    \caption{\textbf{ColEmbed-3B-v2 NDCG@10 by number of annotated pages.}} 
    \label{fig:perf number of pages}
\end{figure}

\paragraph{Visual Content and multi-page queries are hardest for retrievers}
\Cref{fig:content-type-performance} highlights that queries involving visual content like tables or images tend to be more difficult. The Mixed content type scores the lowest, which suggests that integrating information across different modalities within a single page remains a challenge. Additionally, we observe a consistent decline in performance as the number of annotated pages increases (\Cref{fig:perf number of pages}), suggesting that retriever effectiveness decreases
when aggregating information from multiple sources is required.%\vic{think about whether this is normal for ndcg irrespective of query difficulty?} \manu{good point, not sure it's completely comparable even though ndcg is a measure of how perfect the ranking i}.

\paragraph{Cross-language queries degrade performance}
Retrieval performance is 2--3 points higher in mono-lingual settings (\Cref{tab: english-only retrieval} and \Cref{tab: french-only retrieval}) than cross-lingual settings (\Cref{tab:retriever_comparison_updated}), showing that models need to better adapt to these settings. %\anto{Not super interesting}

\newcommand{\plus}[1]{\textcolor{teal}{\tiny \textbf{$\uparrow$#1}}}
\newcommand{\minus}[1]{\textcolor{red}{\tiny \textbf{$\downarrow$#1}}}
\newcommand{\val}[2]{\shortstack{#1 \\[-0.2em] \hspace{1.3em}#2}}
\newcommand{\dline}[2]{\shortstack{#1\\#2}}
\begin{table*}[h]
    \centering
    \small 
    \setlength{\tabcolsep}{3pt} 
    \resizebox{\textwidth}{!}{%
    \begin{tabular}{l ccccccc ccc >{\columncolor{avgcol}}c}
        \toprule
         & \multicolumn{7}{c}{\textbf{English Datasets}} & \multicolumn{3}{c}{\textbf{French Datasets}} & \multicolumn{1}{c}{} \\
        \cmidrule(lr){2-8} \cmidrule(lr){9-11}
        \textbf{Model}& \shortstack{C.S.} & \shortstack{Nucl.} & \shortstack{Fin.} & \shortstack{Phar.} & \shortstack{H.R.} & \shortstack{Ind.} & \shortstack{Tel.} & \shortstack{Phys.} & \shortstack{Ener.} & \shortstack{Fin.} & \textbf{Avg.} \\
        \midrule
        \textit{\textbf{Textual pipeline}} & \multicolumn{11}{c}{}\\
        \multirow[c]{2}{*}{\makecell[l]{Jina-v4\textsuperscript{textual} \\+ zerank-2}} &
        64.3 & 44.3 & 48.4 & 54.9 & 52.8 & 38.4 & 56.3 & 43.6 & 60.1 & 41.3 & 50.4 \\
         &
        \val{\textbf{82.1}}{\plus{17.8}} &
        \val{\textbf{53.5}}{\plus{9.2}} &
        \val{\textbf{69.2}}{\plus{20.8}} &
        \val{\textbf{66.2}}{\plus{11.3}} &
        \val{\textbf{66.5}}{\plus{13.7}} &
        \val{\textbf{53.2}}{\plus{14.8}} &
        \val{\textbf{71.5}}{\plus{15.2}} &
        \val{\textbf{48.2}}{\plus{4.6}} &
        \val{\textbf{71.5}}{\plus{11.4}} &
        \val{\textbf{53.7}}{\plus{12.4}} &
        \val{\textbf{63.6}}{\plus{13.2}} 
        \\
        \midrule
        \textit{\textbf{Visual pipeline}} & \multicolumn{11}{c}{} \\
        \multirow[c]{2}{*}{\makecell[l]{Jina-v4\textsuperscript{visual} \\+ jina-reranker-m0}} & 71.8 & 50.0 & 59.3 & 63.1 & 59.5 & 50.4 & 64.8 & 46.6 & 64.0 & 46.1 & 57.6 \\
        &
        \val{76.7}{\plus{4.9}} & 
        \val{50.8}{\plus{0.8}} & 
        \val{59.2}{\minus{0.1}} & 
        \val{65.4}{\plus{2.3}} & 
        \val{56.0}{\minus{3.5}} & 
        \val{50.9}{\plus{0.5}} & 
        \val{70.8}{\plus{6.0}} & 
        \val{46.9}{\plus{0.3}} & 
        \val{61.7}{\minus{2.3}} & 
        \val{39.8}{\minus{6.3}} & 
        \val{57.8}{\plus{0.2}} \\
        \bottomrule
    \end{tabular}
    }
    \caption{\textbf{Retrieval performance (NDCG@10) of retriever + reranker pipelines.}}
    \label{tab:reranking_scores_stacked}
\end{table*}

\paragraph{Textual rerankers outperform visual ones}

We evaluate the impact of adding a reranker to the textual and visual pipelines of the Jina-v4 retriever. We select zerank-2~\cite{zeroentropy2025} and jina-reranker-m0~\cite{jinareranker2025} as two of the leading textual and visual rerankers to date. Results in Table~\ref{tab:reranking_scores_stacked} reveal a significant disparity in reranking efficacy between modalities. While the visual retriever initially outperforms the textual base, the textual reranker yields substantial gains (+13.2 NDCG@10), enabling the textual pipeline to achieve the highest overall retrieval performance. In contrast, the visual reranker provides only a marginal average improvement of +0.2 and degrades performance in 4 datasets, underscoring the need for better multilingual visual rerankers.

\begin{table*}[h!]
\centering
\small
\resizebox{\textwidth}{!}{%
\begin{tabular}{l l l ccccc ccc >{\columncolor{avgcol}}c c c}
\toprule
& & & \multicolumn{5}{c}{\textbf{English Datasets}} & \multicolumn{3}{c}{\textbf{French Datasets}} & \cellcolor{white}& & \cellcolor{white} \\
\cmidrule(lr){4-8} \cmidrule(lr){9-11}
\makecell[l]{\textbf{Retrieval} \\ \textbf{pipeline}} & \makecell[l]{\textbf{Context} \\ \textbf{modality}} & \makecell[l]{\textbf{Generation} \\ \textbf{model}} & \makecell{C.S.} & \makecell{Fin.} & \makecell{Phar.} & \makecell{H.R.} & \makecell{Ind.} & \makecell{Phys.} & \makecell{Ener.} & \makecell{Fin.} & \makecell{\textbf{Avg.}\\\textbf{Hard}}& \makecell{Avg.\\Easy} & \makecell{Avg.\\Global} \\
\midrule
\multirow{3}{*}{Oracle} & Text & \multirow{3}{*}{Gemini 3 Pro} & 80.9 & 70.2 & 71.4 & \textbf{72.3} & 66.4 & 71.2 & 69.2 & 62.8 & 62.3& 79.3 & 70.6 \\
& Image & & \textbf{86.5} & \textbf{70.6} & \textbf{76.1} & 71.1 & \textbf{68.2} & 74.5 & \textbf{69.8} & \textbf{64.1} & \textbf{64.7}& \textbf{79.7} & \textbf{72.6} \\
& Hybrid & & 86.0 & 68.9 & 73.4 & 70.4 & 65.4 & 69.2 & 69.5 & 62.8 & 63.4& 77.5 & 70.7 \\
\midrule
\makecell[l]{Jina-v4\textsuperscript{text.} + zerank-2} & Text & Gemini 3 Pro & 80.9 & 66.0 & 59.9 & 63.2 & 60.4 & 69.2 & 64.9 & \underline{54.7} & 52.1& 75.5 & 64.9 \\
\midrule
\makecell[l]{Jina-v4\textsuperscript{text.} + zerank-2 \\ \& ColEmbed-3B-v2} & Hybrid & Gemini 3 Pro & 85.1 & 65.0 & 65.9 & 64.8 & 59.4 & 69.9 & 62.7 & 52.8 & \underline{54.7}& 76.6 & 65.7 \\
\midrule
\multirow{6}{*}{ColEmbed-3B-v2} & \multirow{2}{*}{Text} & Gemini 3 Pro & 82.3 & 62.5 & 61.0 & 62.9 & 56.2 & 64.9 & 62.3 & 49.4 & 51.7& 73.0 & 62.7 \\
 & & Kimi K2 & 81.4 & 56.6 & 59.1 & 55.7 & 55.8 & 73.8 & 60.4 & 43.1 & 44.6& 74.3 & 60.7 \\
\cmidrule{2-14}
& \multirow{4}{*}{Image} & Gemini 3 Pro & 83.3 & \underline{67.3} & 62.9 & 65.4 & 57.2 & 67.9 & 64.3 & 47.8 & 54.5& 74.1 & 64.5 \\
 & & Gemini 3 Flash & 80.9 & 64.1 & 63.5 & 63.8 & 55.1 & 68.2 & 63.3 & 47.8 & 50.3& 74.4 & 63.3 \\
& & GPT-5.2 & \underline{\textbf{86.5}} & 59.5 & \underline{68.1} & \underline{66.0} & \underline{61.5} & \underline{\textbf{76.5}} & \underline{66.2} & 49.1 & 54.1& \underline{78.1} & \underline{66.7} \\
& & Qwen3-VL-235B & 86.0 & 59.9 & 64.0 & 60.7 & 57.2 & 71.9 & 59.7 & 44.4 & 51.0& 74.1 & 63.0 \\
\bottomrule
\end{tabular}
}
\caption{\textbf{End-to-end evaluation of final answer generation.} We report the percentage of correct final answers as determined by an LLM judge across the 8 public datasets. "Oracle" rows represent the upper-bound performance using gold-standard contexts. Average Easy and Average Hard denote performance stratified by query difficulty. For each column, the best result is \textbf{bolded} and the best non-oracle result is \underline{underlined}.} % We systematically evaluate the effect of the retrieval pipeline, context modality, and generation model on the generation performance.
\label{tab:end_to_end_eval}
\end{table*}

\subsection{Final Answer Generation} \label{sec:answer results}

We evaluate end-to-end answer quality by providing LLMs and VLMs with queries and their corresponding retrieved pages, examining the effects of retrieval pipeline selection, context modality, and generation model choice (\Cref{tab:end_to_end_eval}). For this evaluation, we use the best-performing textual and visual retrieval pipelines. We additionally establish an upper bound using an \textit{oracle} pipeline that supplies the model with ground-truth annotated pages.

In the \textit{hybrid} configuration, we concatenate the top-5 results from the visual retriever (images) with the top-5 results from the textual retriever (text), without removing duplicates; the retrieval performance is detailed in \Cref{tab:hybrid_comparison}. We also consider a hybrid \textit{oracle} setup, which provides the model with all the ground-truth pages in both modalities.

The correctness of generated answers is assessed against the ground truth final answer by an LLM judge (details in Appendix \ref{appendix:final answer eval}). 
Private datasets are omitted 
to maintain their integrity.

Some benchmark queries involve general knowledge manageable by LLMs without retrieval. To prevent memorization from confounding our assessment of the RAG pipeline, we stratify queries by difficulty based on parametric knowledge. A query is categorized as \textit{easy} if any model in a 6-LLM panel answers it correctly without context; otherwise, it is labeled \textit{hard}. Overall, 48.6\,\% of queries are easy (see~\Cref{tab: easy hard filtering} for details).

\paragraph{Visual context helps generation} With a fixed Gemini 3 Pro generator, image-based context outperforms text-based context on the hard subset by 2.4 and 2.8 percentage points for the oracle and ColEmbed-3B-v2 pipelines, respectively (\Cref{tab:end_to_end_eval}). This confirms that preserving the visual content of document pages provides better grounding for complex answer generation.

\paragraph{Hybrid retrieval yields the best performance on challenging queries} The hybrid pipeline achieves 54.7\,\% accuracy on hard queries, surpassing both the strongest textual (52.1\,\%) and visual (54.5\,\%) baselines. This complementary effect suggests that text and image representations capture different aspects of document content, and their combination can provide more robust evidence for downstream generation. %\que{Not sure about last sentence, seems difficult to back} Conversely, the hybrid oracle (63.4\,\%) trails the image-only upper bound (64.7\,\%), indicating that the redundancy beneficial for retrieval robustness effectively dilutes the signal compared to pure visual grounding.

\paragraph{Hard queries expose the limits of parametric knowledge in current models} Even with oracle context, performance on hard queries lags behind easy queries by more than 10 percentage points. This gap suggests that the multi-step reasoning and long-context synthesis required for difficult queries remain challenging for current models. While the models we evaluate achieve comparable overall scores, their relative ranking may shift when parametric knowledge is less of an advantage, as shown by GPT 5.2 outperforming Gemini 3 Pro on easy queries but trailing on hard ones.

\paragraph{ViDoRe V3 leaves significant room for future retriever improvements} The 10-point gap between the best non-oracle result (54.7\,\%) and the image oracle (64.7\,\%) on hard queries underscores substantial opportunities for improving the retrieval pipeline. Moreover, even with oracle contexts, Gemini 3 Pro performance remains modest, indicating that generation models still struggle to fully exploit the provided information.

\subsection{Visual Grounding} \label{sec:visual grounding results}

Beyond generating correct answers, it is highly desirable for RAG pipelines to identify where in the source documents the answer originates, enabling users to verify the grounding of the query answer. We therefore evaluate the ability of LLMs to generate accurate bounding boxes within their final answer. Among the few LLM families with visual grounding capabilities, we select Qwen3-VL-30B-A3B-Instruct and Gemini 3 Pro for evaluation. For each query, we provide the model with the candidate pages shown to the human annotators and prompt it to answer the query while inserting inline bounding boxes in XML format \texttt{<bboxes image="N"> ... </bboxes>} to delimit relevant content (full instructions in Appendix~\ref{appendix: bbox}).

We use the bounding boxes produced by the human annotators as our ground truth. Since each query may have 1--3 human annotators, we evaluate VLM predictions independently against each annotator using the same zone-based methodology as the inter-annotator consistency analysis (Section~\ref{sec: grounded qa}), and report the highest F1 score. This best-match strategy reflects the inherent subjectivity of evidence selection: annotators may legitimately highlight different regions to support the same answer, and a model should not be penalized for matching any valid interpretation.

\paragraph{Visual grounding lags human performance} 
Inter-annotator agreement on evidence localization reaches an F1 of 0.602, whereas the best-performing models achieve markedly lower scores: 0.089 for Qwen3-VL-30B-A3B-Instruct and 0.065 for Gemini 3 Pro, underlining substantial room for improvement on this task. A page-level analysis (\Cref{tab:bbox_consistency_comparison}) reveals that on pages where humans provided bounding boxes, both models annotated the same page only 16--17\,\% of the time, while 26--27\,\% of human-annotated pages received no model annotation at all, highlighting recall as the primary bottleneck. Per-domain results and qualitative analysis appear in Appendix \ref{appendix: bbox} and \ref{appendix: bbox examples}.

\begin{table}[h!]
\centering
\resizebox{\columnwidth}{!}{%
\begin{tabular}{llcc}
\toprule
\textbf{Category} & \textbf{Outcome} & \textbf{Qwen3-VL-30B-A3B} & \textbf{Gemini 3 Pro} \\
\midrule
\multirow{2}{*}{\textbf{Agreement}} 
  & Both annotated & 17\,\% & 16\,\% \\
  & Neither annotated  & 46\,\% & 49\,\% \\
\addlinespace
\multirow{2}{*}{\textbf{Discrepancy}} 
  & Model only   & 10\,\% & 7\,\% \\
  & Human only & 26\,\% & 27\,\% \\
\bottomrule
\end{tabular}
}
\caption{\textbf{Page-level bounding box agreement between models and human annotators.} Each page is classified by whether the model and human both annotated it, both left it unannotated, or only one provided annotations.}
\label{tab:bbox_consistency_comparison}
\end{table}

\section{Conclusion}

This work introduces ViDoRe V3, a human-annotated RAG benchmark that evaluates cross-lingual retrieval, final answer generation, and visual grounding on large industry-relevant document corpora. We design a human-in-the-loop annotation methodology, deployed in a 12,000-hour annotation campaign, that produces diverse realistic queries paired with relevant pages, bounding boxes, and reference answers. Evaluating state-of-the-art RAG pipelines, we find that visual retrievers outperform textual ones, late interaction and textual reranking yield substantial gains, and visual context improves answer generation quality. Looking ahead, ViDoRe V3 highlights several concrete research directions for practical multimodal RAG. Retriever models still struggle on cross-lingual and open-ended queries requiring visual interpretation, while VLMs need improvement in answer generation from multi-page contexts as well as accurate visual grounding. By providing a rigorous framework for evaluating these limitations, ViDoRe V3 serves as a catalyst for the development of more robust, intelligent document understanding models.

\section*{Limitations}

\paragraph{Language coverage}
While our benchmark is multilingual, it is restricted to English and French source documents and queries in 6 high-resource Western European languages. Future iterations of the benchmark should include a more diverse set of language families and non-Latin scripts to mitigate this bias.

\paragraph{Document distribution bias} 
Our benchmark focuses on publicly available long-form document corpora, representing one specific mode of existing document distribution. For example, enterprise RAG may need to handle a wider variety of document types, often in private repositories, that include noisy, short-form types such as emails, support tickets, or scanned handwritten notes that are not represented in our source documents.

\paragraph{Human annotation}
Annotations for open-ended reasoning and visual grounding inherently contain a degree of subjectivity. We acknowledge that for complex exploratory queries, multiple valid retrieval paths and answer formulations may exist outside of our annotated ground truths.

\section*{Ethical considerations}

\paragraph{Annotator Welfare and Compensation.} Human annotation was conducted by the creators of the benchmark and a single external annotation vendor. Multiple established vendors were evaluated with respect to the annotation protocol and relevant ethical considerations, and one vendor was selected based on demonstrated compliance with these criteria. Annotators were recruited from the vendor's existing workforce in accordance with the demographic requirements described in the Annotator Pool and Selection section (\Cref{appendix: annotator pool}) and were compensated at rates designed to provide fair pay based on geographic location and required skill sets.  The data were curated such that annotators were not exposed to harmful or offensive content during the annotation process. The use of human annotators was limited to standard annotation and verification tasks for benchmark construction and did not constitute human-subjects research; accordingly, the data collection protocol was determined to be exempt from formal ethics review.

\paragraph{Data Licensing and Privacy.} All documents included in the benchmark were manually selected from governmental, educational, and enterprise websites that met open license criteria. The annotations were collected in order not to contain any private or personally identifiable information and are GDPR-compliant. The benchmark is released under a commercially permissive license to facilitate broad research adoption while respecting the intellectual property rights of original document creators.

\paragraph{Linguistic and Geographic Bias.} We acknowledge that our benchmark is restricted to English and French source documents and queries in 6 high-resource Western European languages. This limitation may inadvertently favor RAG systems optimized for these languages and does not reflect the full diversity of practical document retrieval scenarios globally. We encourage future work to extend evaluation to underrepresented language families and non-Latin scripts.

\paragraph{Environmental Impact.} The creation of this benchmark required substantial computational resources for VLM pre-filtering, synthetic query generation, and model evaluation. We report these costs to promote transparency: approximately 12,000 hours of human annotation effort and extensive GPU compute for model inference across our evaluation suite. Specifically, the compute totaled 3,000 hours on NVIDIA H100 GPUs on a low emission energy grid, with an estimated environmental impact of 200 kg $\text{CO}_2\text{e}$.

\section*{Acknowledgments}

This work was conducted with contributions from NVIDIA. We thank all the people that allowed this work to happen, in particular Eric Tramel, Benedikt Schifferer, Mengyao Xu and Radek Osmulski, Erin Potter and Hannah Brandon. Crucially, we thank the dedicated team of annotators for their essential efforts.\\
It was carried out within the framework of the LIAGORA "LabCom", a joint laboratory supported by the French National Research Agency (ANR) and established between ILLUIN Technology and the MICS laboratory of CentraleSupelec. The benchmark was partially created using HPC resources from IDRIS with grant AD011016393.

\section*{Detailed Contributions}

\paragraph{Benchmark Design} Loison, Macé, Edy, Moreira and Liu designed the benchmark.

\paragraph{Data and Annotation} Loison and Macé developed the synthetic data generation pipeline. Loison generated the queries, while Macé predicted links between queries and pages. Loison, Macé, and Balough defined annotation guidelines; Balough coordinated the annotation campaign. Macé and Edy managed final answer merging. Loison, Macé, Edy, Xing, and Balough reviewed the final annotations.

\paragraph{Evaluation} Macé, Edy and Loison conceptualized the evaluations. Macé and Loison worked on retrieval evaluation, with Moreira focusing on the evaluation of ColEmbed models. Edy led the end-to-end evaluation, reranking analysis, and visualization. Macé and Edy integrated the results into the MTEB leaderboard. Xing led bounding box evaluations and result analysis.

\paragraph{Writing and Supervision} The manuscript was written by Loison, Macé, Xing, and Edy. Senior supervision and strategic guidance were provided by Xing, Faysse, Liu, Hudelot, and Viaud, with Faysse closely advising on project direction and planning.

\bibliography{custom}

@misc{rteb2025,
  author = {Liu, Frank and Enevoldsen, Kenneth and Solomatin, Roman and Chung, Isaac and Aarsen, Tom and Fődi, Zoltán},
  url = {https://huggingface.co/blog/rteb},
  title = {Introducing RTEB: A New Standard for Retrieval Evaluation},
  year = {2025},
}

@article{liquidai2025lfm2,
 title={LFM2 Technical Report},
 author={{Liquid AI}},
 journal={arXiv preprint arXiv:2511.23404},
 year={2025}
}

@misc{takehi2025fantasticsmallretrieverstrain,
      title={Fantastic (small) Retrievers and How to Train Them: mxbai-edge-colbert-v0 Tech Report}, 
      author={Rikiya Takehi and Benjamin Clavié and Sean Lee and Aamir Shakir},
      year={2025},
      eprint={2510.14880},
      archivePrefix={arXiv},
      primaryClass={cs.IR},
      url={https://arxiv.org/abs/2510.14880}, 
}

@article{teiletche2025modernvbert,
  title={ModernVBERT: Towards Smaller Visual Document Retrievers},
  author={Teiletche, Paul and Mac{\'e}, Quentin and Conti, Max and Loison, Antonio and Viaud, Gautier and Colombo, Pierre and Faysse, Manuel},
  journal={arXiv preprint arXiv:2510.01149},
  year={2025}
}

@misc{nomicembedmultimodal2025,
  title={Nomic Embed Multimodal: Interleaved Text, Image, and Screenshots for Visual Document Retrieval},
  author={{Nomic Team}},
  year={2025},
  publisher={Nomic AI},
  url={https://nomic.ai/blog/posts/nomic-embed-multimodal},
}

@misc{xu2025llamanemoretrievercolembedtopperforming,
      title={Llama Nemoretriever Colembed: Top-Performing Text-Image Retrieval Model}, 
      author={Mengyao Xu and Gabriel Moreira and Ronay Ak and Radek Osmulski and Yauhen Babakhin and Zhiding Yu and Benedikt Schifferer and Even Oldridge},
      year={2025},
      eprint={2507.05513},
      archivePrefix={arXiv},
      primaryClass={cs.CV},
      url={https://arxiv.org/abs/2507.05513}, 
}

@misc{bm25s,
      title={BM25S: Orders of magnitude faster lexical search via eager sparse scoring}, 
      author={Xing Han Lù},
      year={2024},
      eprint={2407.03618},
      archivePrefix={arXiv},
      primaryClass={cs.IR},
      url={https://arxiv.org/abs/2407.03618}, 
}

@article{wang2025vidorag,
  title={Vidorag: Visual document retrieval-augmented generation via dynamic iterative reasoning agents},
  author={Wang, Qiuchen and Ding, Ruixue and Chen, Zehui and Wu, Weiqi and Wang, Shihang and Xie, Pengjun and Zhao, Feng},
  journal={arXiv preprint arXiv:2502.18017},
  year={2025}
}

@article{wasserman2025real,
  title={REAL-MM-RAG: A Real-World Multi-Modal Retrieval Benchmark},
  author={Wasserman, Navve and Pony, Roi and Naparstek, Oshri and Goldfarb, Adi Raz and Schwartz, Eli and Barzelay, Udi and Karlinsky, Leonid},
  journal={arXiv preprint arXiv:2502.12342},
  year={2025}
}

@article{su2024bright,
  title={Bright: A realistic and challenging benchmark for reasoning-intensive retrieval},
  author={Su, Hongjin and Yen, Howard and Xia, Mengzhou and Shi, Weijia and Muennighoff, Niklas and Wang, Han-yu and Liu, Haisu and Shi, Quan and Siegel, Zachary S and Tang, Michael and others},
  journal={arXiv preprint arXiv:2407.12883},
  year={2024}
}

@article{wang2024charxiv,
  title={Charxiv: Charting gaps in realistic chart understanding in multimodal llms},
  author={Wang, Zirui and Xia, Mengzhou and He, Luxi and Chen, Howard and Liu, Yitao and Zhu, Richard and Liang, Kaiqu and Wu, Xindi and Liu, Haotian and Malladi, Sadhika and others},
  journal={Advances in Neural Information Processing Systems},
  volume={37},
  pages={113569--113697},
  year={2024}
}

@inproceedings{mathew2021docvqa,
  title={Docvqa: A dataset for vqa on document images},
  author={Mathew, Minesh and Karatzas, Dimosthenis and Jawahar, CV},
  booktitle={Proceedings of the IEEE/CVF winter conference on applications of computer vision},
  pages={2200--2209},
  year={2021}
}

@misc{jinareranker2025,
  author = {{Jina AI}},
  title = {jina-reranker-m0: Multilingual Multimodal Document Reranker},
  url = {https://jina.ai/news/jina-reranker-m0-multilingual-multimodal-document-reranker},
  year = {2025},
  note = {Accessed: 2025-12-22}
}

@misc{zeroentropy2025,
  author = {{Zero Entropy}},
  title = {Introducing zerank-2},
  url = {https://www.zeroentropy.dev/articles/zerank-2-advanced-instruction-following-multilingual-reranker},
  year = {2025},
  note = {Accessed: 2025-12-22}
}

@Manual{nvidiaocr,
  title = {NVIDIA Ingest: An accelerated pipeline for document ingestion},
  author = {{NVIDIA Ingest Development Team}},
  year = {2024},
  url = {https://github.com/NVIDIA/nv-ingest},
}

@article{cho2024m3docrag,
  title={M3docrag: Multi-modal retrieval is what you need for multi-page multi-document understanding},
  author={Cho, Jaemin and Mahata, Debanjan and Irsoy, Ozan and He, Yujie and Bansal, Mohit},
  journal={arXiv preprint arXiv:2411.04952},
  year={2024}
}

@inproceedings{muennighoff2023mteb,
  title={Mteb: Massive text embedding benchmark},
  author={Muennighoff, Niklas and Tazi, Nouamane and Magne, Lo{\"\i}c and Reimers, Nils},
  booktitle={Proceedings of the 17th Conference of the European Chapter of the Association for Computational Linguistics},
  pages={2014--2037},
  year={2023}
}

@inproceedings{van2023document,
  title={Document understanding dataset and evaluation (dude)},
  author={Van Landeghem, Jordy and Tito, Rub{\`e}n and Borchmann, {\L}ukasz and Pietruszka, Micha{\l} and Joziak, Pawel and Powalski, Rafal and Jurkiewicz, Dawid and Coustaty, Micka{\"e}l and Anckaert, Bertrand and Valveny, Ernest and others},
  booktitle={Proceedings of the IEEE/CVF International Conference on Computer Vision},
  pages={19528--19540},
  year={2023}
}

@misc{faysse2025colpaliefficientdocumentretrieval,
      title={ColPali: Efficient Document Retrieval with Vision Language Models}, 
      author={Manuel Faysse and Hugues Sibille and Tony Wu and Bilel Omrani and Gautier Viaud and Céline Hudelot and Pierre Colombo},
      year={2025},
      eprint={2407.01449},
      archivePrefix={arXiv},
      primaryClass={cs.IR},
      url={https://arxiv.org/abs/2407.01449}, 
}

@misc{marafioti2025smolvlmredefiningsmallefficient,
      title={SmolVLM: Redefining small and efficient multimodal models}, 
      author={Andrés Marafioti and Orr Zohar and Miquel Farré and Merve Noyan and Elie Bakouch and Pedro Cuenca and Cyril Zakka and Loubna Ben Allal and Anton Lozhkov and Nouamane Tazi and Vaibhav Srivastav and Joshua Lochner and Hugo Larcher and Mathieu Morlon and Lewis Tunstall and Leandro von Werra and Thomas Wolf},
      year={2025},
      eprint={2504.05299},
      archivePrefix={arXiv},
      primaryClass={cs.AI},
      url={https://arxiv.org/abs/2504.05299}, 
}

@inproceedings{zhu2022towards,
  title={Towards complex document understanding by discrete reasoning},
  author={Zhu, Fengbin and Lei, Wenqiang and Feng, Fuli and Wang, Chao and Zhang, Haozhou and Chua, Tat-Seng},
  booktitle={Proceedings of the 30th ACM International Conference on Multimedia},
  pages={4857--4866},
  year={2022}
}

@misc{chen_bge_2024,
	title = {{BGE} {M3}-{Embedding}: {Multi}-{Lingual}, {Multi}-{Functionality}, {Multi}-{Granularity} {Text} {Embeddings} {Through} {Self}-{Knowledge} {Distillation}},
	copyright = {Creative Commons Attribution 4.0 International},
	shorttitle = {{BGE} {M3}-{Embedding}},
	url = {https://arxiv.org/abs/2402.03216},
	doi = {10.48550/ARXIV.2402.03216},
	urldate = {2024-06-04},
	publisher = {arXiv},
	author = {Chen, Jianlv and Xiao, Shitao and Zhang, Peitian and Luo, Kun and Lian, Defu and Liu, Zheng},
	year = {2024},
	note = {Version Number: 3},
}

@misc{GTE-ModernColBERT,
title={GTE-ModernColBERT},
author={Chaffin, Antoine},
url={https://huggingface.co/lightonai/GTE-ModernColBERT-v1},
year={2025}
}

@misc{ma2024unifyingmultimodalretrievaldocument,
      title={Unifying Multimodal Retrieval via Document Screenshot Embedding}, 
      author={Xueguang Ma and Sheng-Chieh Lin and Minghan Li and Wenhu Chen and Jimmy Lin},
      year={2024},
      eprint={2406.11251},
      archivePrefix={arXiv},
      primaryClass={cs.IR},
      url={https://arxiv.org/abs/2406.11251}, 
}

@misc{thakur2025freshstackbuildingrealisticbenchmarks,
      title={FreshStack: Building Realistic Benchmarks for Evaluating Retrieval on Technical Documents}, 
      author={Nandan Thakur and Jimmy Lin and Sam Havens and Michael Carbin and Omar Khattab and Andrew Drozdov},
      year={2025},
      eprint={2504.13128},
      archivePrefix={arXiv},
      primaryClass={cs.IR},
      url={https://arxiv.org/abs/2504.13128}, 
}

@misc{macé2025vidorebenchmarkv2raising,
      title={ViDoRe Benchmark V2: Raising the Bar for Visual Retrieval}, 
      author={Quentin Macé and António Loison and Manuel Faysse},
      year={2025},
      eprint={2505.17166},
      archivePrefix={arXiv},
      primaryClass={cs.IR},
      url={https://arxiv.org/abs/2505.17166}, 
}

@misc{qwen3technicalreport,
      title={Qwen3 Technical Report}, 
      author={{Qwen Team}},
      year={2025},
      eprint={2505.09388},
      archivePrefix={arXiv},
      primaryClass={cs.CL},
      url={https://arxiv.org/abs/2505.09388}, 
}

@misc{günther2025jinaembeddingsv4universalembeddingsmultimodal,
      title={jina-embeddings-v4: Universal Embeddings for Multimodal Multilingual Retrieval}, 
      author={Michael Günther and Saba Sturua and Mohammad Kalim Akram and Isabelle Mohr and Andrei Ungureanu and Sedigheh Eslami and Scott Martens and Bo Wang and Nan Wang and Han Xiao},
      year={2025},
      eprint={2506.18902},
      archivePrefix={arXiv},
      primaryClass={cs.AI},
      url={https://arxiv.org/abs/2506.18902}, 
}

@misc{yu2025visragvisionbasedretrievalaugmentedgeneration,
      title={VisRAG: Vision-based Retrieval-augmented Generation on Multi-modality Documents}, 
      author={Shi Yu and Chaoyue Tang and Bokai Xu and Junbo Cui and Junhao Ran and Yukun Yan and Zhenghao Liu and Shuo Wang and Xu Han and Zhiyuan Liu and Maosong Sun},
      year={2025},
      eprint={2410.10594},
      archivePrefix={arXiv},
      primaryClass={cs.IR},
      url={https://arxiv.org/abs/2410.10594}, 
}

@misc{mathew2021infographicvqa,
      title={InfographicVQA}, 
      author={Minesh Mathew and Viraj Bagal and Rubèn Pérez Tito and Dimosthenis Karatzas and Ernest Valveny and C. V Jawahar},
      year={2021},
      eprint={2104.12756},
      archivePrefix={arXiv},
      primaryClass={cs.CV},
      url={https://arxiv.org/abs/2104.12756}, 
}

@article{qwen25vl,
  title={Qwen2.5-VL Technical Report},
  author={Bai, Shuai and Chen, Keqin and Liu, Xuejing and Wang, Jialin and Ge, Wenbin and Song, Sibo and Dang, Kai and Wang, Peng and Wang, Shijie and Tang, Jun and Zhong, Humen and Zhu, Yuanzhi and Yang, Mingkun and Li, Zhaohai and Wan, Jianqiang and Wang, Pengfei and Ding, Wei and Fu, Zheren and Xu, Yiheng and Ye, Jiabo and Zhang, Xi and Xie, Tianbao and Cheng, Zesen and Zhang, Hang and Yang, Zhibo and Xu, Haiyang and Lin, Junyang},
  journal={arXiv preprint arXiv:2502.13923},
  year={2025}
}

@misc{peng2025unidocbenchunifiedbenchmarkdocumentcentric,
      title={UNIDOC-BENCH: A Unified Benchmark for Document-Centric Multimodal RAG}, 
      author={Xiangyu Peng and Can Qin and Zeyuan Chen and Ran Xu and Caiming Xiong and Chien-Sheng Wu},
      year={2025},
      eprint={2510.03663},
      archivePrefix={arXiv},
      primaryClass={cs.CL},
      url={https://arxiv.org/abs/2510.03663}, 
}

@article{auer2024docling,
  title={Docling technical report},
  author={Auer, Christoph and Lysak, Maksym and Nassar, Ahmed and Dolfi, Michele and Livathinos, Nikolaos and Vagenas, Panos and Ramis, Cesar Berrospi and Omenetti, Matteo and Lindlbauer, Fabian and Dinkla, Kasper and others},
  journal={arXiv preprint arXiv:2408.09869},
  year={2024}
}

@article{qwen3embedding,
  title={Qwen3 Embedding: Advancing Text Embedding and Reranking Through Foundation Models},
  author={Zhang, Yanzhao and Li, Mingxin and Long, Dingkun and Zhang, Xin and Lin, Huan and Yang, Baosong and Xie, Pengjun and Yang, An and Liu, Dayiheng and Lin, Junyang and Huang, Fei and Zhou, Jingren},
  journal={arXiv preprint arXiv:2506.05176},
  year={2025}
}

@misc{mcinnes2020umapuniformmanifoldapproximation,
      title={UMAP: Uniform Manifold Approximation and Projection for Dimension Reduction}, 
      author={Leland McInnes and John Healy and James Melville},
      year={2020},
      eprint={1802.03426},
      archivePrefix={arXiv},
      primaryClass={stat.ML},
      url={https://arxiv.org/abs/1802.03426}, 
}

@inproceedings{campello2013density,
  title={Density-based clustering based on hierarchical density estimates},
  author={Campello, Ricardo JGB and Moulavi, Davoud and Sander, J{\"o}rg},
  booktitle={Pacific-Asia conference on knowledge discovery and data mining},
  pages={160--172},
  year={2013},
  organization={Springer}
}

@misc{lewis2021retrievalaugmentedgenerationknowledgeintensivenlp,
      title={Retrieval-Augmented Generation for Knowledge-Intensive NLP Tasks}, 
      author={Patrick Lewis and Ethan Perez and Aleksandra Piktus and Fabio Petroni and Vladimir Karpukhin and Naman Goyal and Heinrich Küttler and Mike Lewis and Wen-tau Yih and Tim Rocktäschel and Sebastian Riedel and Douwe Kiela},
      year={2021},
      eprint={2005.11401},
      archivePrefix={arXiv},
      primaryClass={cs.CL},
      url={https://arxiv.org/abs/2005.11401}, 
}

@misc{gao2024retrievalaugmentedgenerationlargelanguage,
      title={Retrieval-Augmented Generation for Large Language Models: A Survey}, 
      author={Yunfan Gao and Yun Xiong and Xinyu Gao and Kangxiang Jia and Jinliu Pan and Yuxi Bi and Yi Dai and Jiawei Sun and Meng Wang and Haofen Wang},
      year={2024},
      eprint={2312.10997},
      archivePrefix={arXiv},
      primaryClass={cs.CL},
      url={https://arxiv.org/abs/2312.10997}, 
}

@misc{fan2024surveyragmeetingllms,
      title={A Survey on RAG Meeting LLMs: Towards Retrieval-Augmented Large Language Models}, 
      author={Wenqi Fan and Yujuan Ding and Liangbo Ning and Shijie Wang and Hengyun Li and Dawei Yin and Tat-Seng Chua and Qing Li},
      year={2024},
      eprint={2405.06211},
      archivePrefix={arXiv},
      primaryClass={cs.CL},
      url={https://arxiv.org/abs/2405.06211}, 
}

@misc{abootorabi2025askmodalitycomprehensivesurvey,
      title={Ask in Any Modality: A Comprehensive Survey on Multimodal Retrieval-Augmented Generation}, 
      author={Mohammad Mahdi Abootorabi and Amirhosein Zobeiri and Mahdi Dehghani and Mohammadali Mohammadkhani and Bardia Mohammadi and Omid Ghahroodi and Mahdieh Soleymani Baghshah and Ehsaneddin Asgari},
      year={2025},
      eprint={2502.08826},
      archivePrefix={arXiv},
      primaryClass={cs.CL},
      url={https://arxiv.org/abs/2502.08826}, 
}

@misc{gao2023enablinglargelanguagemodels,
      title={Enabling Large Language Models to Generate Text with Citations}, 
      author={Tianyu Gao and Howard Yen and Jiatong Yu and Danqi Chen},
      year={2023},
      eprint={2305.14627},
      archivePrefix={arXiv},
      primaryClass={cs.CL},
      url={https://arxiv.org/abs/2305.14627}, 
}

@misc{ma2024visaretrievalaugmentedgeneration,
      title={VISA: Retrieval Augmented Generation with Visual Source Attribution}, 
      author={Xueguang Ma and Shengyao Zhuang and Bevan Koopman and Guido Zuccon and Wenhu Chen and Jimmy Lin},
      year={2024},
      eprint={2412.14457},
      archivePrefix={arXiv},
      primaryClass={cs.IR},
      url={https://arxiv.org/abs/2412.14457}, 
}

@misc{cho2024m3docragmultimodalretrievalneed,
      title={M3DocRAG: Multi-modal Retrieval is What You Need for Multi-page Multi-document Understanding}, 
      author={Jaemin Cho and Debanjan Mahata and Ozan Irsoy and Yujie He and Mohit Bansal},
      year={2024},
      eprint={2411.04952},
      archivePrefix={arXiv},
      primaryClass={cs.CV},
      url={https://arxiv.org/abs/2411.04952}, 
}

@misc{tang2024multihopragbenchmarkingretrievalaugmentedgeneration,
      title={MultiHop-RAG: Benchmarking Retrieval-Augmented Generation for Multi-Hop Queries}, 
      author={Yixuan Tang and Yi Yang},
      year={2024},
      eprint={2401.15391},
      archivePrefix={arXiv},
      primaryClass={cs.CL},
      url={https://arxiv.org/abs/2401.15391}, 
}

@misc{yu2025bboxdocvqalargescale,
      title={BBox DocVQA: A Large Scale Bounding Box Grounded Dataset for Enhancing Reasoning in Document Visual Question Answer}, 
      author={Wenhan Yu and Wang Chen and Guanqiang Qi and Weikang Li and Yang Li and Lei Sha and Deguo Xia and Jizhou Huang},
      year={2025},
      eprint={2511.15090},
      archivePrefix={arXiv},
      primaryClass={cs.DB},
      url={https://arxiv.org/abs/2511.15090}, 
}

@misc{nemo-data-designer,
  author = {{NeMo Data Designer Team}},
  title = {NeMo Data Designer: A framework for generating synthetic data from scratch or based on your own seed data},
  howpublished = {\url{https://github.com/NVIDIA-NeMo/DataDesigner}},
  year = {2025},
  note = {GitHub Repository},
}

@misc{conti2025contextgoldgoldpassage,
      title={Context is Gold to find the Gold Passage: Evaluating and Training Contextual Document Embeddings}, 
      author={Max Conti and Manuel Faysse and Gautier Viaud and Antoine Bosselut and Céline Hudelot and Pierre Colombo},
      year={2025},
      eprint={2505.24782},
      archivePrefix={arXiv},
      primaryClass={cs.IR},
      url={https://arxiv.org/abs/2505.24782}, 
}

\clearpage
\appendix

\onecolumn

\section{Dataset examples}

\begin{figure*}[h]
    \centering
    \includegraphics[width=0.92\textwidth]{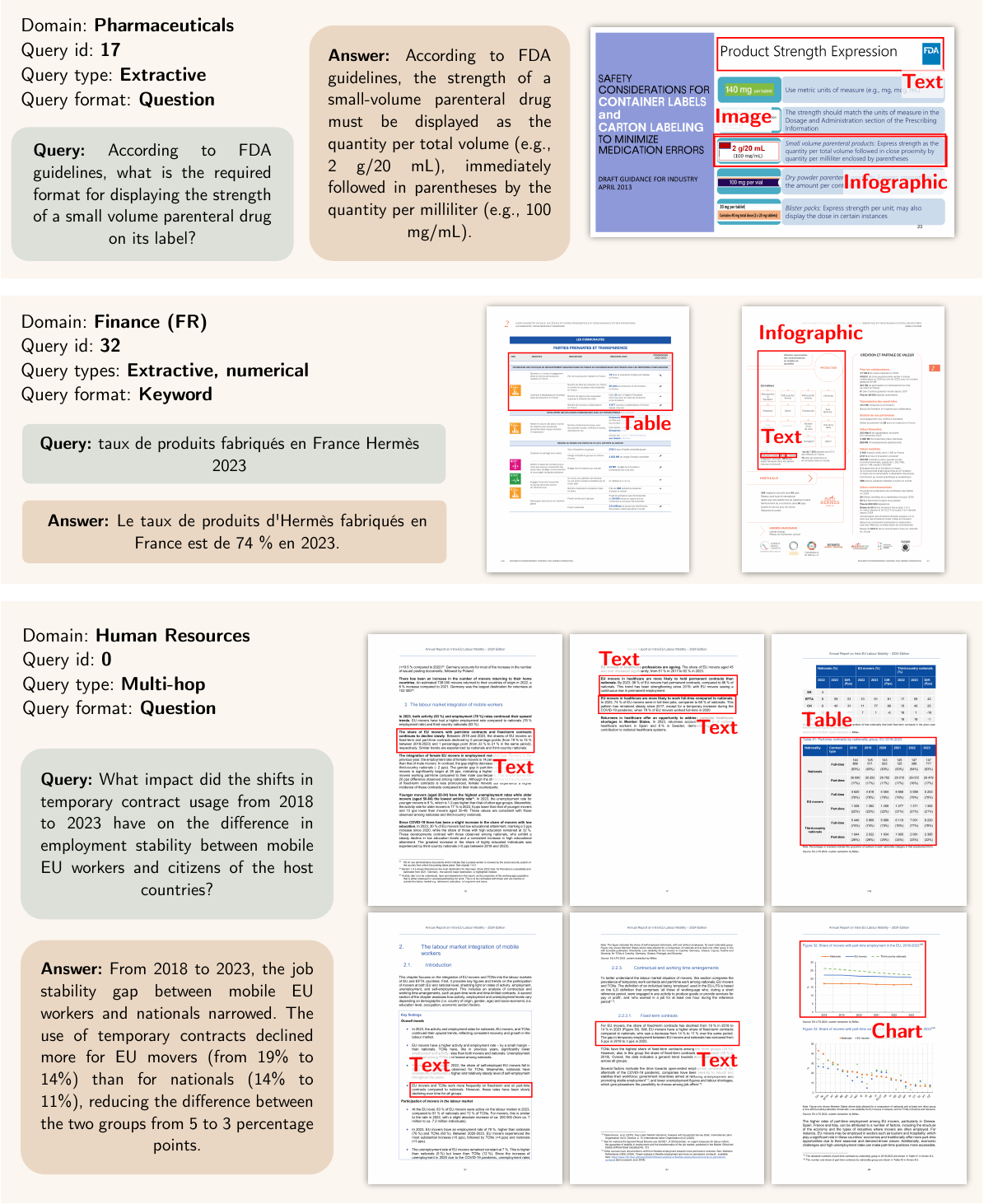}
    \caption{\textbf{Examples from the ViDoRe V3 datasets.} Featuring varied query types and visually rich document formats across multiple domains, the benchmark captures the complexity of real-world retrieval scenarios.}
\end{figure*}

\twocolumn

\section{Supplementary benchmark details}

\paragraph{Domains} \Cref{tab:datasets} details the type of documents used in each corpus as well as several statistics.

\paragraph{Query type and format descriptions}

\begin{figure*}[t!]
    \centering
    \includegraphics[width=\textwidth]{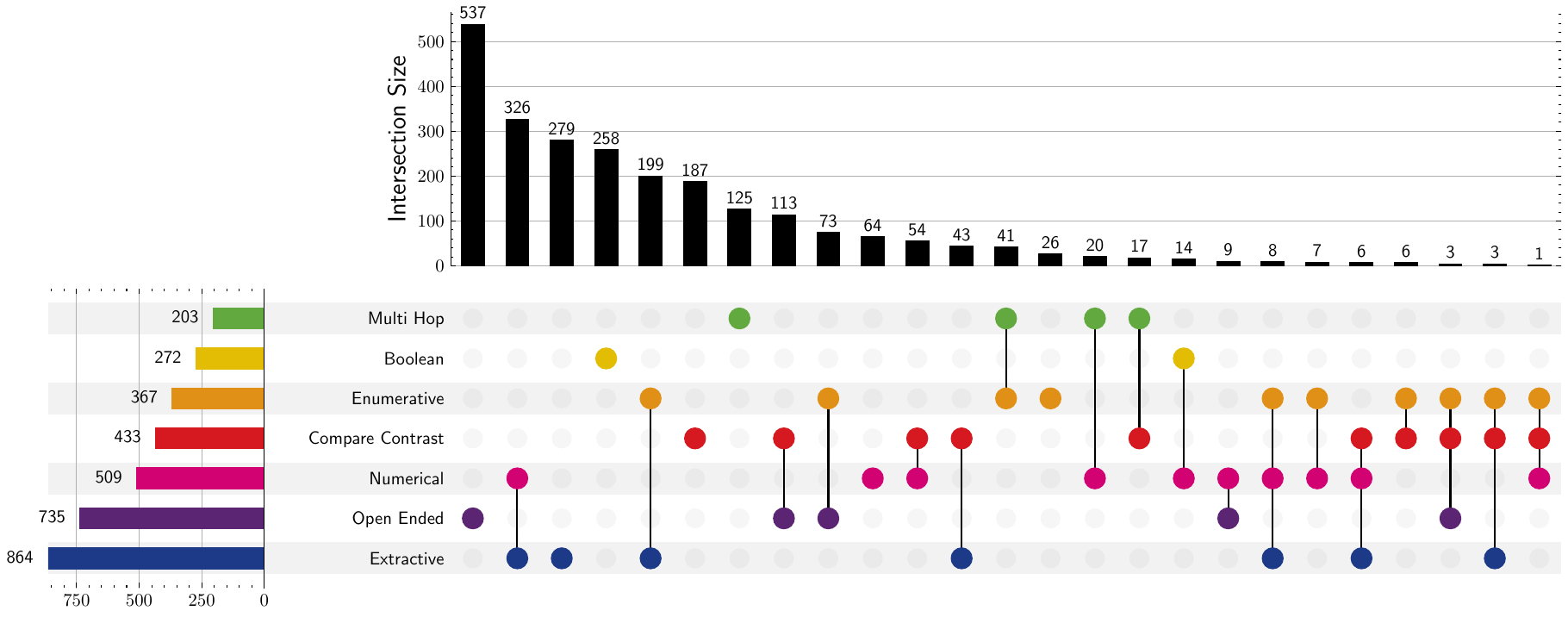}
    \caption{\textbf{UpSet plot illustrating the distribution and intersection of query types in ViDoRe V3.} The horizontal bars on the left display the total count of queries for each individual type. The vertical bars at the top represent the frequency of specific combinations (intersections), as indicated by the connected dots in the matrix below. While \textit{Extractive} queries are the most prevalent overall, \textit{Open Ended} queries form the dominant unique category. Complex dependencies are evident in the frequent intersection of \textit{Enumerative} and \textit{Extractive types}, indicating a substantial subset of queries requiring list-based fact retrieval.} 
    \label{fig:upset plot}
\end{figure*}

\Cref{tab:query_taxonomy} describes the types and formats of the queries, while \Cref{fig:upset plot} gives details about query type intersection frequency.

\paragraph{Query type by generation method}

\begin{figure*}[t!]
    \centering
    \includegraphics[clip, width=\textwidth]{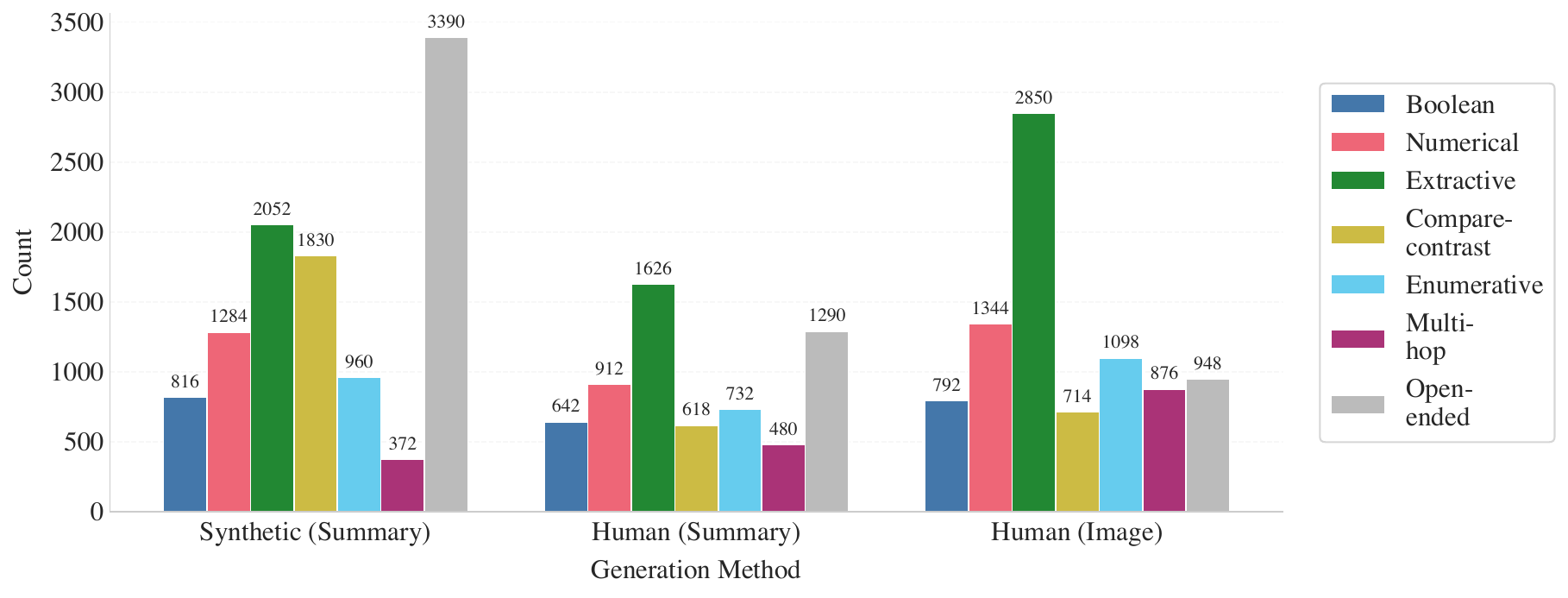}
    \caption{\textbf{Query type distribution by generation method.}}
    \label{fig:query type by generation method}
\end{figure*}

Query type distributions by generation method (Figure~\ref{fig:query type by generation method}) confirm that open-ended queries dominate synthetic queries as the synthetic pipeline attributed more weight to this type, while extractive queries dominate human-image queries since they are more naturally chosen by annotators.

\section{Annotator pool and training details}
\label{appendix: annotator pool}
\paragraph{Annotator Pool and Selection.} Annotation was conducted by a curated pool of 76 annotators who were selected based on having: (1) a bachelor's degree or higher in the relevant domain, (2) professional experience in the domain, (3) native-level language proficiency as required by task, and (4) prior experience with RAG, retrieval, or VQA annotation projects. Quality control was performed by 13 senior annotators with enhanced domain knowledge and extensive annotation experience, with project oversight provided by data leads with multiple years of experience in human data generation.

\paragraph{Training and Pilot Phase.} The annotation process began with a comprehensive onboarding phase where annotators received task-specific training using gold-standard examples. For each domain, a pilot of several hundred tasks was conducted with 100\% quality control coverage and multiple annotators per task. During this phase, data leads and the research team continuously evaluated annotations, provided clarifications, and refined guidelines. Inter-annotator agreement and time-per-task baselines were calculated to establish ongoing evaluation benchmarks. The pilot concluded upon validation of both data quality and guideline effectiveness.

\section{Supplementary agreement metrics}
\label{appendix: relevance agreement}

Pages were pre-filtered by a VLM before human annotation; as most pages shown to annotators were likely relevant, this created a skewed class distribution. This prevalence imbalance causes traditional chance-corrected metrics like Krippendorff's Alpha to appear paradoxically low even when annotators genuinely agree, as inflated expected chance agreement penalizes the score. To address this, we report 2 complementary metrics: Krippendorff's Alpha (ordinal) as the standard measure and Gwet's AC2 which remains stable under prevalence skew. Overall, annotators achieved $\alpha = 0.469$, AC2 $= 0.760$. The divergence between Alpha and AC2/Weighted Agreement is expected given the pre-filtered data and confirms substantial agreement despite the skewed distribution.

\section{Supplementary retrieval details}
\label{appendix:supp retrieval}
\paragraph{Retriever model reference}

Table~\ref{tab:model_metadata} lists the retriever models evaluated in this work, along with their HuggingFace model names and citations.

\paragraph{Monolingual performance} \label{sec:english retrievers}

Tables~\ref{tab: english-only retrieval} and \ref{tab: french-only retrieval} present the monolingual performance of our models, where retrieval is conducted using language-matched queries and documents for English and French, respectively.

\begin{table*}[t!]
\centering
\small % Options: \small, \footnotesize (smaller), or \scriptsize (smallest)
\begin{tabularx}{\textwidth}{lXX} % 'l' for auto-width category, 'X' for wrapping columns
\toprule
\textbf{Category} & \textbf{Definition} & \textbf{Example} \\ \midrule
\multicolumn{3}{c}{\textit{\textbf{Query Types}}} \\ \midrule
\textbf{Open-ended} & Seeks explanatory or descriptive information that requires synthesis. & What drives the rise in women's workforce involvement in EU nations? \\
\textbf{Extractive} & Requires the retrieval of a specific piece of information. & Bank of America preferred stock MM dividend rate \\
\textbf{Compare Contrast} & Mandates a comparison between multiple entities or data points. & Explain the factors contributing to the reduction in R2R rates for ANDAs. \\
\textbf{Boolean} & Poses a yes/no question necessitating multi-step reasoning. & Did JPMorganChase execute more than half of its planned repurchase program? \\
\textbf{Numerical} & Asks for a specific quantitative value that must be derived or calculated. & percentage increase in Morgan Stanley revenue from 2023 to 2024 \\
\textbf{Multi-hop} & Requires integrating information from multiple sections or sources. & Summarize the steps involved in error reporting in ISMP's MERP. \\
\textbf{Enumerative} & Requests a list of all instances sharing a common property. & Specify the ISCO codes used to define domestic workers in the EU. \\ \midrule
\multicolumn{3}{c}{\textit{\textbf{Query Formats}}} \\ \midrule
\textbf{Question} & An interrogative sentence. & What was Citigroup's net interest margin in 2024? \\
\textbf{Keyword} & A non-verbal phrase or set of terms. & female employment rate European Union 2023 \\
\textbf{Instruction} & A directive specifying a task. & Identify the use case of a drill point gauge. \\ \bottomrule
\end{tabularx}
\caption{Taxonomy of Query Types and Formats.}
\label{tab:query_taxonomy}
\end{table*}

\begin{table*}[t]
\small
\centering
\setlength{\tabcolsep}{4pt}
\renewcommand{\arraystretch}{1.15}

\begin{tabularx}{\textwidth}{
    >{\RaggedRight\arraybackslash}X
    >{\RaggedRight\arraybackslash}X
    >{\RaggedRight\arraybackslash}p{4.2cm}
    c
    c
    c
    c
    >{\RaggedRight\arraybackslash}X
}
\toprule
\textbf{Corpus} &
\textbf{Domain(s)} &
\textbf{Description} &
\textbf{Lang.} &
\textbf{\# Docs} &
\textbf{\# Pages} &
\textbf{\# Queries$^{*}$} &
\textbf{Main modalities} \\
\midrule

\textbf{U.S. Public Company Annual Reports} &
Finance-EN &
Consists of 6 10-K annual reports from major U.S. financial institutions for the fiscal year ended December 31, 2024. &
en & 6 & 2942 & 309 &
Text, Table \\
\addlinespace

\textbf{Computer Science Textbooks} &
Computer Science / Education &
Consists of two open-source, peer-reviewed textbooks from OpenStax covering foundational topics in computer science, Python, and data science. &
en & 2 & 1360 & 215 &
Books \\
\addlinespace

\textbf{FDA Reports} &
Pharmaceuticals &
Consists of FDA presentations and Springer books (2016--2023) covering regulatory policies, drug development, and public health initiatives. &
en & 52 & 2313 & 364 &
Slides, Books \\
\addlinespace

\textbf{HR Reports from EU} &
HR &
Includes recent European Commission reports and papers on EU labour markets, social development, and employment policies. &
en & 14 & 1110 & 318 &
Reports \\
\addlinespace

\textbf{USAF Technical Orders} &
Industrial Maintenance &
Comprises U.S. military technical orders and manuals for aircraft maintenance, safety procedures, and material handling, revised through 2025. &
en & 27 & 5244 & 283 &
Manuals \\
\addlinespace

\textbf{French Physics Lectures} &
Physics &
A collection of educational materials offering an interdisciplinary exploration of modern physics and complexity science. &
fr & 42 & 1674 & 302 &
Slides \\
\addlinespace

\textbf{French Public Company Annual Reports} &
Finance-FR &
Contains the 2023--2024 annual reports of major French luxury companies (Dior, Hermès, Kering, L'Oréal, LVMH). &
fr & 5 & 2384 & 320 &
Reports \\
\addlinespace

\textbf{French Governmental Energy Reports} &
Energy &
Gathers official documents from French public agencies on energy, economic, and environmental issues in France. &
fr & 42 & 2229 & 308 &
Reports, Slides \\
\bottomrule
\end{tabularx}

\caption{Description of ViDoRe V3 public corpora. $^{*}$Number of queries is without translations}
\label{tab:datasets}
\end{table*}

    % Nuclear Energy & 
    % Contains regulatory documents from 2004-2006 concerning the Vermont Yankee Nuclear Power Station's proposed power uprate. &
    % en & 
    % 48 & 
    % 2814 & 
    
    % Reports, Email \\
    % \addlinespace

    % \textbf{Internet} \newline \textbf{Protocols} & 
    % Telecom & 
    % A collection of IETF RFCs and standards documents defining core internet protocols, network management, and communication standards. & 
    % en & 
    % 123 & 
    % 4627 & 
    %  & 
    % Reports \\
    % \addlinespace

\begin{table}[H]
\centering

\resizebox{\columnwidth}{!}{%
\begin{tabular}{lccc}
\toprule
\textbf{Dataset} & \textbf{$\alpha$ (ord)} & \textbf{Gwet's AC2} \\ %& \textbf{Wgt. Agree} \\
\midrule
Computer science& 0.467 & 0.809 \\%& 0.849 \\
Energy & 0.463 & 0.714 \\%& 0.805 \\
Finance (EN) & 0.514 & 0.798 \\%& 0.832 \\
Finance (FR) & 0.320 & 0.736 \\%& 0.801 \\
H.R. & 0.413 & 0.793 \\%& 0.840 \\
Industrial Maintenance & 0.496 & 0.740 \\%& 0.803 \\
Telecom & 0.464 & 0.772 \\%& 0.833 \\
Nuclear & 0.389 & 0.794 \\%& 0.837 \\
Pharma & 0.478 & 0.755 \\%& 0.831 \\
Physics & 0.213 & 0.334 \\%& 0.668 \\
\midrule
\textbf{Overall} & \textbf{0.469} & \textbf{0.760} \\%& \textbf{0.815} \\
\bottomrule
\end{tabular}
}
\caption{Inter-annotator agreement for relevance ratings by dataset.}
\label{tab:iaa}
\end{table}

\begin{table*}[ht]
\centering
\begin{tabular}{lll}
\toprule
\textbf{Model alias} & \textbf{Full model name} & \textbf{Reference} \\ \midrule
Qwen3-8B & Qwen3-Embedding-8B & \citet{qwen3embedding} \\
Jina-v4 & jina-embeddings-v4 & \citet{günther2025jinaembeddingsv4universalembeddingsmultimodal} \\
Qwen3-0.6B & Qwen3-Embedding-0.6B & \citet{qwen3embedding} \\
LFM2-350M & LFM2-ColBERT-350M & \citet{liquidai2025lfm2} \\
BGE-M3 & BGE-M3 & \citet{chen_bge_2024} \\
BM25S & BM25S & \citet{bm25s} \\
ColEmbed-3B-v2 & llama-nemoretriever-colembed-3b-v2 & \citet{xu2025llamanemoretrievercolembedtopperforming} \\ % asked Gabriel about this
ColNomic-7B & colnomic-embed-multimodal-7b & \citet{nomicembedmultimodal2025} \\
ColEmbed-3B & llama-nemoretriever-colembed-3b-v1 & \citet{xu2025llamanemoretrievercolembedtopperforming} \\
ColNomic-3B & colnomic-embed-multimodal-3b & \citet{nomicembedmultimodal2025} \\
ColEmbed-1B & llama-nemoretriever-colembed-1b-v1 & \citet{xu2025llamanemoretrievercolembedtopperforming} \\
ColQwen2.5 & colqwen2.5-v0.2 & \citet{faysse2025colpaliefficientdocumentretrieval} \\
Nomic-7B & nomic-embed-multimodal-7b & \citet{nomicembedmultimodal2025} \\
ColQwen2 & colqwen2-v1.0 & \citet{faysse2025colpaliefficientdocumentretrieval} \\
Nomic-3B & nomic-embed-multimodal-7b & \citet{nomicembedmultimodal2025} \\
ColPali & colpali-v1.3 & \citet{faysse2025colpaliefficientdocumentretrieval} \\
Mxbai Edge 32M & mxbai-edge-colbert-v0-32m & \citet{takehi2025fantasticsmallretrieverstrain} \\
GTE-ModernColBERT & GTE-ModernColBERT-v1 & \citet{GTE-ModernColBERT} \\
ColModernVBERT & colmodernvbert & \citet{teiletche2025modernvbert} \\
ColSmol-256M & colSmol-256M & \citet{marafioti2025smolvlmredefiningsmallefficient} \\
\bottomrule
\end{tabular}
\caption{\textbf{Retriever reference table.} Model aliases used in Tables~\ref{tab:retriever_comparison_updated},~\ref{tab: english-only retrieval}, and~\ref{tab: french-only retrieval} are mapped to their HuggingFace model name and citation.}
\label{tab:model_metadata}
\end{table*}

\begin{table*}[h!]
\centering
\begin{tabular}{l c c c c c c c c >{\columncolor{avgcol}}c}
\toprule
\textbf{Model} & Size (B) & C.S. & Nucl. & Fin. & Phar. & H.R. & Ind. & Tele. & \textbf{Average} \\
\midrule
\multicolumn{9}{l}{\textit{\textbf{Textual Retrievers}}} \\
Jina-v4	& 3 & 67.3	& \textbf{48.2} & \textbf{56.5}&  59.0& \textbf{58.8} & 	45.8	& 61.0	& \textbf{56.7}\\
Qwen3-8B$^\bigstar$	& 8 & \textbf{73.5}	& 42.2	 & 54.8	& 62.4	& 52.3	& 45.3	& \textbf{66.0}	& 56.6 \\
LFM2-350M & 0.35 & 70.6 & 45.4 & 48.3 & 62.1 & 53.2 & \textbf{47.9} & 63.8 & 55.9\\
Mxbai Edge 32M & 0.03 & 68.0 & 44.4 & 48.2 & \textbf{62.5} & 52.7 & 47.1 & 61.9 & 55.0 \\
BM25S	& - & 64.7&	45.9&	49.9&	56.9&	49.6&	45.6	&58.3	&53.0\\
Qwen3-0.6B$^\bigstar$& 0.6 &	70.5	& 39.7	&51.5	& 57.4	& 46.2	&42.4	&59.7	&52.3\\
GTE-ModernColBERT & 0.15 & 63.6 & 41.7 & 39.8 & 62.0 & 46.2 & 44.6 & 59.7 & 51.1\\
BGE-M3$^\bigstar$	& 0.57 &63.6&	34.3	& 43.9	&54.7	&45.3	&39.0	&54.3	&47.9\\
\midrule
\multicolumn{9}{l}{\textit{\textbf{Visual Retrievers}}} \\
ColEmbed-3B-v2 & 3 & \textbf{78.6} & 52.9 & 69.1 & \textbf{67.6} & \textbf{65.4} & 56.8 & \textbf{71.7} & \textbf{66.0} \\
ColEmbed-3B & 3 & 77.8 & \textbf{53.4} & \textbf{69.5 }& 66.9 & 64.9 & \textbf{57.0} & 69.4 & 65.6 \\
ColEmbed-1B & 1 & 75.5 & 52.2 & 67.0 & 66.2 & 64.5 & 56.1 & 68.7 & 64.3 \\
Jina-v4 & 3 & 74.2 & 52.4 & 66.1 & 65.2 & 64.6 & 55.9 & 68.7 & 63.9 \\
ColNomic-7B & 7 & 78.2 & 48.2 & 63.1 & 64.6 & 62.9 & 54.2 & 69.6 & 63.0 \\
ColNomic-3B & 3 & 75.5 & 45.5 & 63.0 & 63.7 & 62.6 & 52.8 & 68.6 & 61.7 \\
ColQwen2.5 & 3 & 75.2 & 42.9 & 61.2 & 60.9 & 59.2 & 49.4 & 65.3 & 59.2 \\
Nomic-7B$^\bigstar$ & 7 & 70.9 & 42.3 & 57.6 & 63.8 & 55.9 & 48.5 & 62.0 & 57.3 \\
ColQwen2 & 2 & 73.5 & 44.1 & 50.9 & 58.1 & 54.7 & 49.8 & 63.2 & 56.3 \\
ColPali & 7 & 72.5 & 38.1 & 43.3 & 57.7 & 53.3 & 47.0 & 59.2 & 53.0 \\
Nomic-3B$^\bigstar$ & 3 & 62.1 & 37.2 & 53.3 & 59.2 & 51.9 & 41.1 & 57.2 & 51.7 \\
ColModernVBERT & 0.25 & 59.7 & 42.0 & 50.4 & 56.6 & 47.0 & 43.9 & 55.2 & 50.7 \\
ColSmol-256M & 0.25 & 57.4 & 36.5 & 47.7 & 51.4 & 46.0 & 38.5 & 47.5 & 46.4 \\
\bottomrule
\end{tabular}
\caption{\textbf{English-only retrieval performance (NDCG@10).} $\bigstar$: single-vector models. Results are computed on the English queries of the English datasets.}
\label{tab: english-only retrieval}
\end{table*}

\begin{table*}[h!]
\small
\centering
\begin{tabular}{l c c c c >{\columncolor{avgcol}}c}
\toprule
\textbf{Model} & Size (B) & Phys. & Ener. & Fin. & \textbf{Average} \\
\midrule
\multicolumn{5}{l}{\textit{\textbf{Textual Retrievers}}} \\
Jina-v4	& 3 & 44.0	& \textbf{63.4}	& \textbf{44.8} & \textbf{50.7}\\
Qwen3-8B$^\bigstar$	& 8 &\textbf{45.8}	& 60.2	& 37.6& 47.9\\
Qwen3-0.6B$^\bigstar$	& 0.6 & 43.8	& 54.9	& 28.5& 42.4\\
BGE-M3$^\bigstar$ &0.57	& 38.3	& 53.1	& 28.4&39.9\\
BM25S	& - & 39.8	& 57.4	& 35.9 & 44.4\\
\midrule
\multicolumn{5}{l}{\textit{\textbf{Visual Retrievers}}} \\
ColEmbed-3B-v2 & 3 & 48.2 & 67.5 & 48.2 & \textbf{54.6}\\
ColNomic-7b	& 7 & \textbf{48.5} & 67.0	& 47.9 & 54.5\\
ColNomic-3b	& 3 & \textbf{48.5} &	\textbf{67.9} &	46.8 & 54.4\\
Jina-v4	& 3 & 46.8	& 66.7 & 48.6 & 54.0 \\
ColEmbed-3B & 	3 & 46.6 &	66.3 &	\textbf{48.9} & 53.9\\
ColEmbed-1B & 1 & 44.7 &	64.6 &	47.8 & 52.4\\
ColQwen2.5 &	3 & 47.8	&62.3	&43.6& 51.2\\
Nomic-7B$^\bigstar$ &	7 & 45.6&	61.6&	41.3 & 49.5\\
ColQwen2	& 3 & 43.9	&55.6	&26.5& 42.0\\
Nomic-3B$^\bigstar$	& 3 & 43.6	&56.4	&34.4 & 44.8\\
ColPali	& 7 & 43.2	&50.5	&23.6 & 39.1 \\
\bottomrule
\end{tabular}
\caption{\textbf{French-only retrieval performance (NDCG@10).} $\bigstar$: single-vector models. Results are computed on the French queries of the French datasets.}
\label{tab: french-only retrieval}
\end{table*}

\paragraph{Additional Retrieval Modality Performances} To evaluate the hybrid retrieval setup, we use the multimodal Jina-v4 model to generate separate visual and textual rankings. We then construct a hybrid retrieval set by merging the top-5 results from each modality and removing duplicates. Because this set-union operation does not preserve a strict ranking order, we report the unranked F1 score. As shown in \Cref{tab:hybrid_comparison}, the hybrid approach consistently outperforms single-modality baselines.

\begin{table*}[t!]
\centering
\small
\begin{tabular}{l c c c c c c c c c c >{\columncolor{avgcol}}c}
\toprule
 & \multicolumn{7}{c}{\textbf{English Datasets}} & \multicolumn{3}{c}{\textbf{French Datasets}} & \multicolumn{1}{c}{} \\
\cmidrule(lr){2-8} \cmidrule(lr){9-11}
\textbf{Modality} & \shortstack{C.S.} & \shortstack{Nucl.} & \shortstack{Fin.} & \shortstack{Phar.} & \shortstack{H.R.} & \shortstack{Ind.} & \shortstack{Tel.} & \shortstack{Phys.} & \shortstack{Ener.} & \shortstack{Fin.} & \textbf{Avg.} \\
\midrule
Visual	& 39.4	& 25.5	& 28.4	& 27.5	& 30.0	& 21.4	& 31.4	&\textbf{26.6}	& 25.2	& 22.9	& 27.8 \\
Textual	& 35.4	& 23.1&	24.5	&24.2	&27.4&	16.5&	29.0&	25.8 &	23.9	&20.4&	25.0 \\
Hybrid	&\textbf{43.0}	&\textbf{27.7}	&\textbf{30.9}	&\textbf{29.7}	&\textbf{32.6}	&\textbf{22.2}	&\textbf{35.5}&	26.5	&\textbf{29.8}	&\textbf{24.3}	&\textbf{30.2} \\
\midrule
\textit{Avg. \# Pages for hybrid}	&\textit{6.96}	& \textit{7.38}	&\textit{7.77}&\textit{7.40}&	\textit{7.29}&	\textit{7.77}	&\textit{7.09} &	\textit{7.26}&	\textit{6.97}	&\textit{7.61}&	\textit{7.35} \\
\bottomrule
\end{tabular}
\caption{\textbf{Performance comparison of retrieval modalities (F1@10) on Jina-v4.} Evaluation is performed using the multimodal retriever Jina-v4. The Hybrid method combines the top-5 visual and top-5 textual matches, subsequently removing duplicates. The final row reports the average number of unique pages remaining in the hybrid set. The Hybrid setup constantly outperforms both textual and visual retrieval.}
\label{tab:hybrid_comparison}
\end{table*}

\section{ColEmbed-3B-v2 performance breakdown}\label{appendix: colembed-v2-perf-breakdown}

\Cref{tab:perf language} details the retrieval scores of ColEmbed-3B-v2 by query language, highlighting small performance variations by language.

\paragraph{Performance by number of annotated pages}

\begin{figure}[t]
    \centering
    \includegraphics[width=0.9\columnwidth]{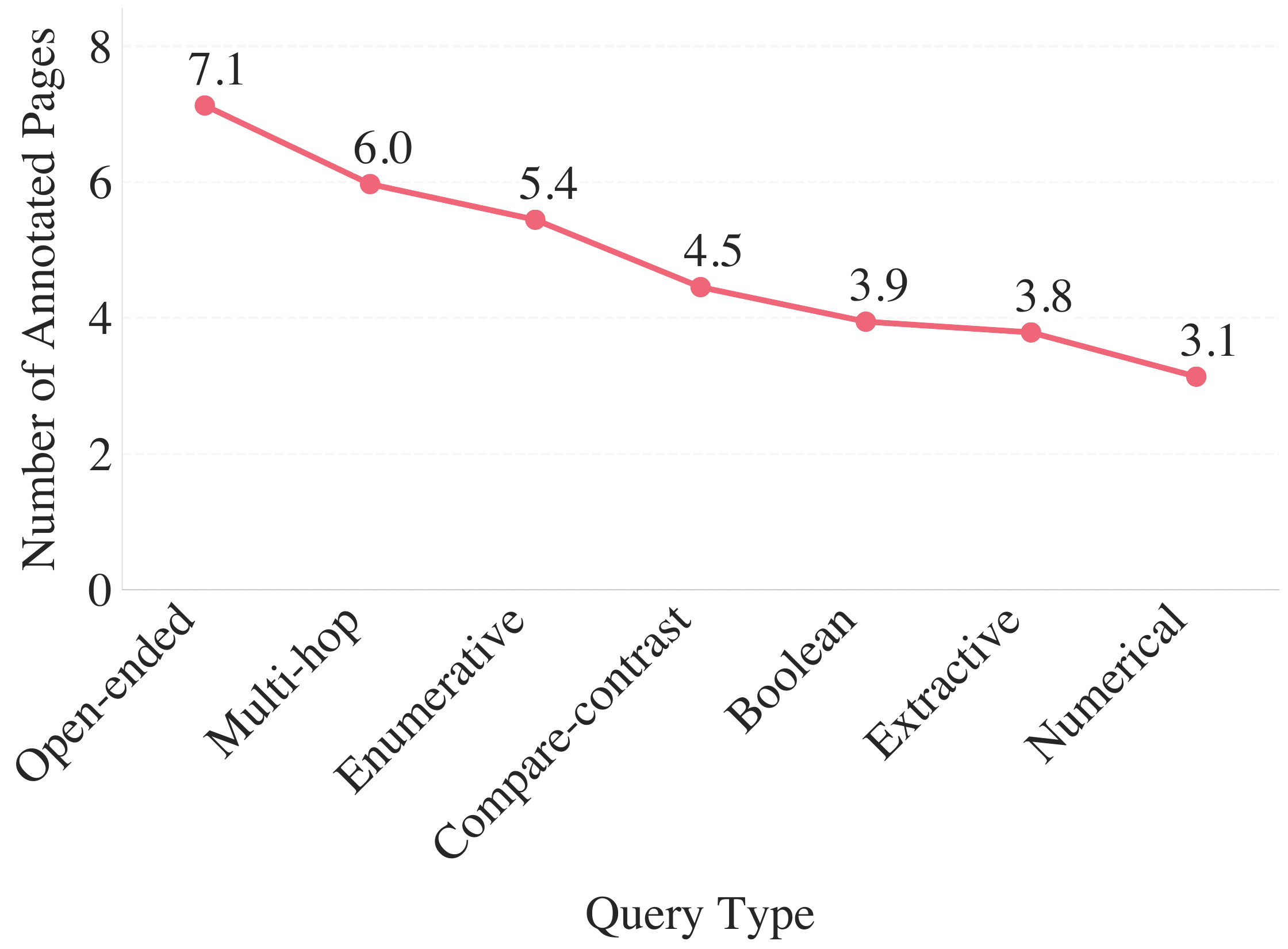}
    \caption{\textbf{Average number of annotated pages by query type.}} 
    \label{fig:num pages per query type}
\end{figure}

As seen in Figure~\ref{fig:perf number of pages}, performance drops with the number of annotated pages. However, a potential confounding factor is the correlation between query type and the number of annotated pages, since more complex query types also have higher number of annotated pages (Figure~\ref{fig:num pages per query type}). We perform a stratified regression analysis to isolate these two effects.

We model NDCG@10 as a linear function of the number of annotated pages ($P$) stratified by query type. For each of the 7 query types, we fit an ordinary least squares regression:
\begin{equation*}
    NDCG@10 = a \cdot P + b + \epsilon.
\end{equation*}

\begin{figure}[h!]
    \centering
    \includegraphics[width=0.8\columnwidth]{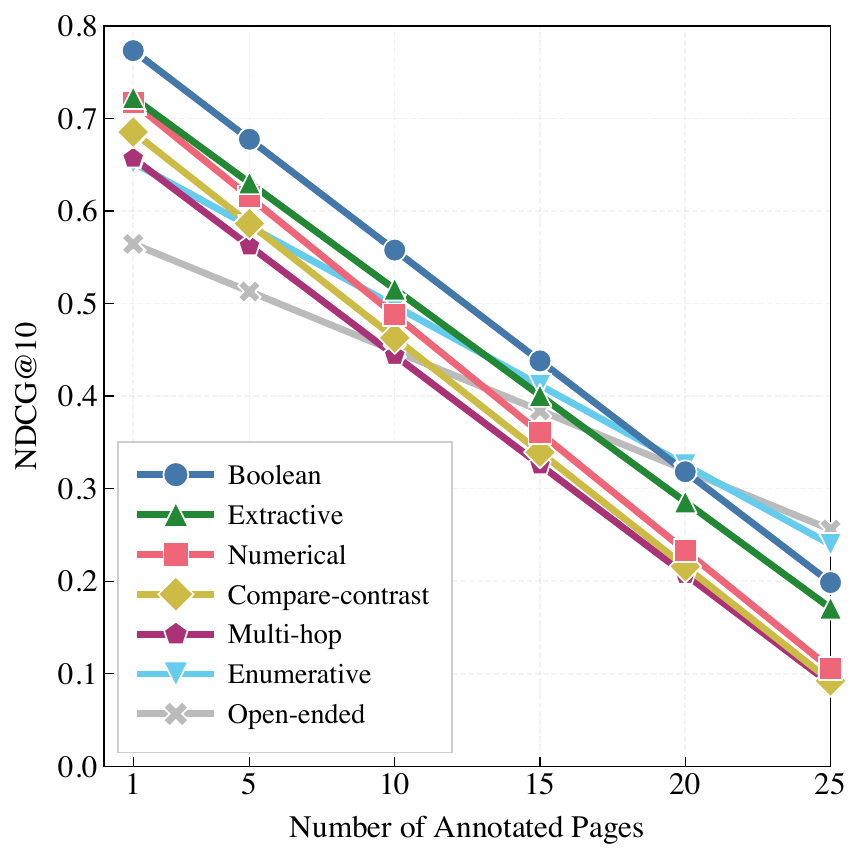}
    \caption{\textbf{ColEmbed-3B-v2 NDCG@10 by number of annotated pages and query type.}} 
    \label{fig:perf content num pages}
\end{figure}

Results in Figure~\ref{fig:perf content num pages} and Table~\ref{tab:num pages content type decay} reveal that all query types suffer a significant performance penalty as the number of annotated pages increases. Slope values are nearly uniform ($a \approx -0.024$), suggesting a similar drop in retrieval accuracy across most query types. The open-ended and enumerative types are the two exceptions: despite having the lowest NDCG@10 for low page counts, they also have the shallowest slope, which suggests that retrieval success on these queries is constrained by the model's fundamental difficulty in synthesizing multiple relevant sources rather than the volume of relevant context.

\begin{table}[!h]
\small
\centering
\begin{tabular}{lc}
\toprule
\textbf{Query Language} & \textbf{NDCG@10}\\
\midrule
English & 60.8 \\
French & 59.8 \\
Portuguese & 59.6 \\
Spanish & 59.6 \\
Italian & 59.1 \\
German & 57.9 \\
\bottomrule
\end{tabular}
\caption{\textbf{ColEmbed-3B-v2 NDCG@10 by query language.}}
\label{tab:perf language}
\end{table}

\begin{table}[h!]
\centering
\small
\begin{tabular}{lccc}
\toprule
\textbf{Query Type} & Slope $a$ & Intercept $b$ & $R^2$ \\
\midrule
Boolean           & -0.0239 & 0.797 & 0.101 \\
Numerical         & -0.0255 & 0.742 & 0.059 \\
Extractive        & -0.0230 & 0.745 & 0.084 \\
Compare-contrast  & -0.0247 & 0.710 & 0.117 \\
Enumerative       & -0.0172 & 0.669 & 0.080 \\
Multi-hop         & -0.0237 & 0.680 & 0.114 \\
Open-ended        & -0.0129 & 0.577 & 0.057 \\
\bottomrule
\end{tabular}
\caption{\textbf{Linear regression analysis of NDCG@10 decay with number of annotated pages, by query type.} The slope $a$ represents performance sensitivity to retrieval context size, while the intercept $b$ represents intrinsic difficulty at minimum context size.}
\label{tab:num pages content type decay}
\end{table}

\paragraph{Performance by Query Generation Source}

Table~\ref{tab:perf_query_source} reports NDCG@10 of ColEmbed-3B-v2 stratified by query type and generation source. Queries generated with full page image access (Human+Image) consistently yield the highest scores, as annotators can anchor queries directly to visible content. Synthetic and human blind-contextual queries (SDG+Summary and Human+Summary) score similarly, with a mean difference of 0.044 NDCG@10, confirming that the synthetic pipeline does not introduce a simplicity bias.

\begin{table}[h!]
\centering
\small
\begin{tabular}{lccc}
\toprule
\textbf{Query Type} & \begin{tabular}[c]{@{}c@{}}\textbf{SDG+}\\ \textbf{Sum.}\end{tabular} & \begin{tabular}[c]{@{}c@{}}\textbf{Human+}\\ \textbf{Sum.}\end{tabular} & \begin{tabular}[c]{@{}c@{}}\textbf{Human+}\\ \textbf{Img.}\end{tabular} \\
\midrule
Boolean          & 0.615 & 0.692 & 0.801 \\
Compare-Contrast & 0.573 & 0.514 & 0.743 \\
Open-Ended       & 0.447 & 0.459 & 0.657 \\
Extractive       & 0.568 & 0.623 & 0.743 \\
Enumerative      & 0.449 & 0.559 & 0.697 \\
Numerical        & 0.592 & 0.659 & 0.732 \\
Multi-Hop        & 0.398 & 0.378 & 0.688 \\
\midrule
\textbf{Mean}    & \textbf{0.521} & \textbf{0.565} & \textbf{0.726} \\
\bottomrule
\end{tabular}
\caption{\textbf{ColEmbed-3B-v2 NDCG@10 by query type and generation source.} \textbf{SDG+Sum.:} synthetic blind-contextual;\textbf{ Human+Sum:}: human blind-contextual; \textbf{Human+Img:}: human with full page image access.}
\label{tab:perf_query_source}
\end{table}

\paragraph{Performance by content type}

\begin{figure}[h!]
\centering
\includegraphics[width=\columnwidth]{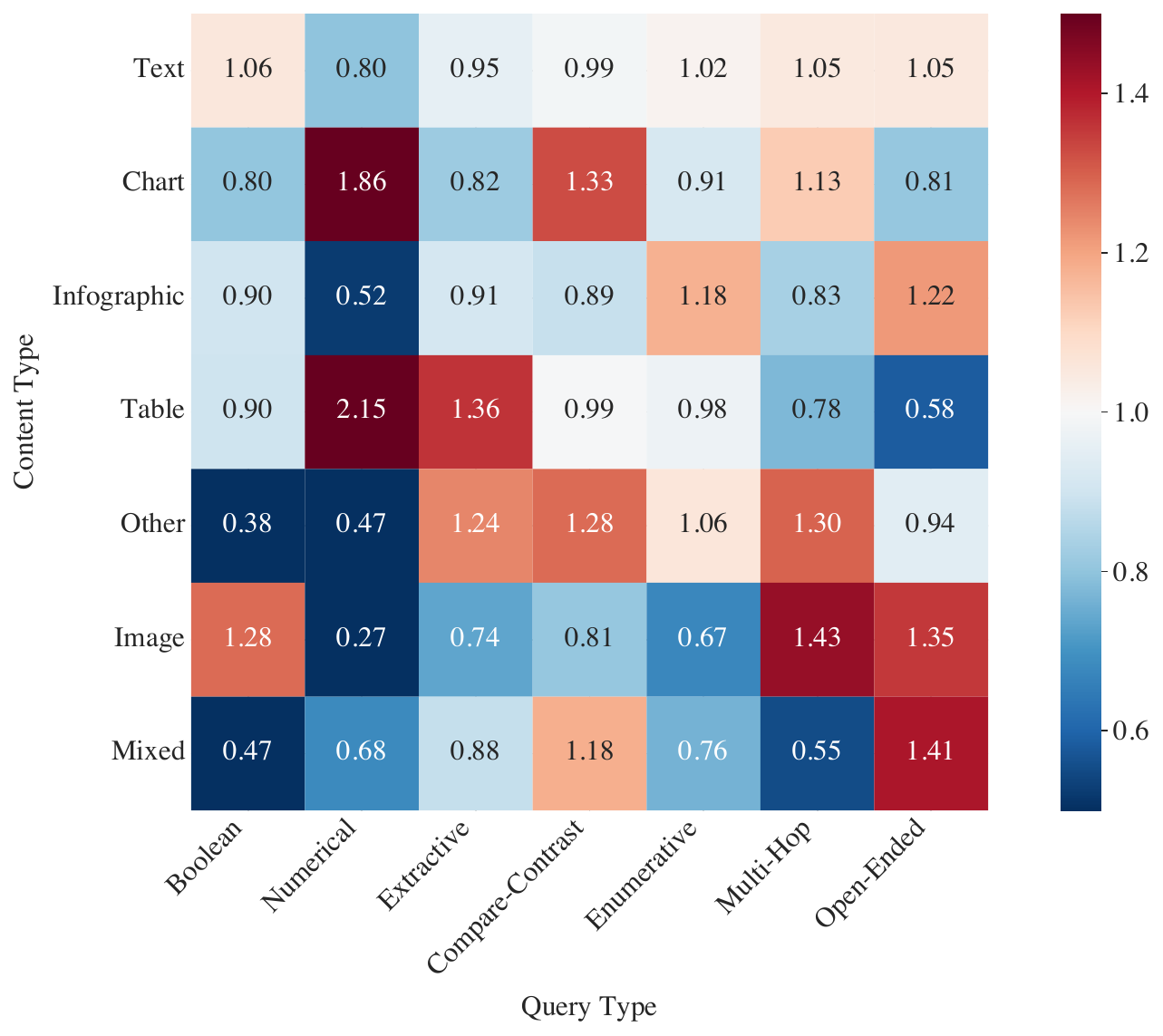}
\caption{\textbf{Lift of query types by content type.} Each cell shows the ratio of observed query type frequency to baseline frequency for a given content type. Values >1 indicate over-representation (e.g., tables appear 2.15× more in numerical queries than expected), while values <1 indicate under-representation.}
\label{fig:lift content type}
\end{figure}

\begin{figure}[h!]
\centering
\includegraphics[width=\columnwidth]{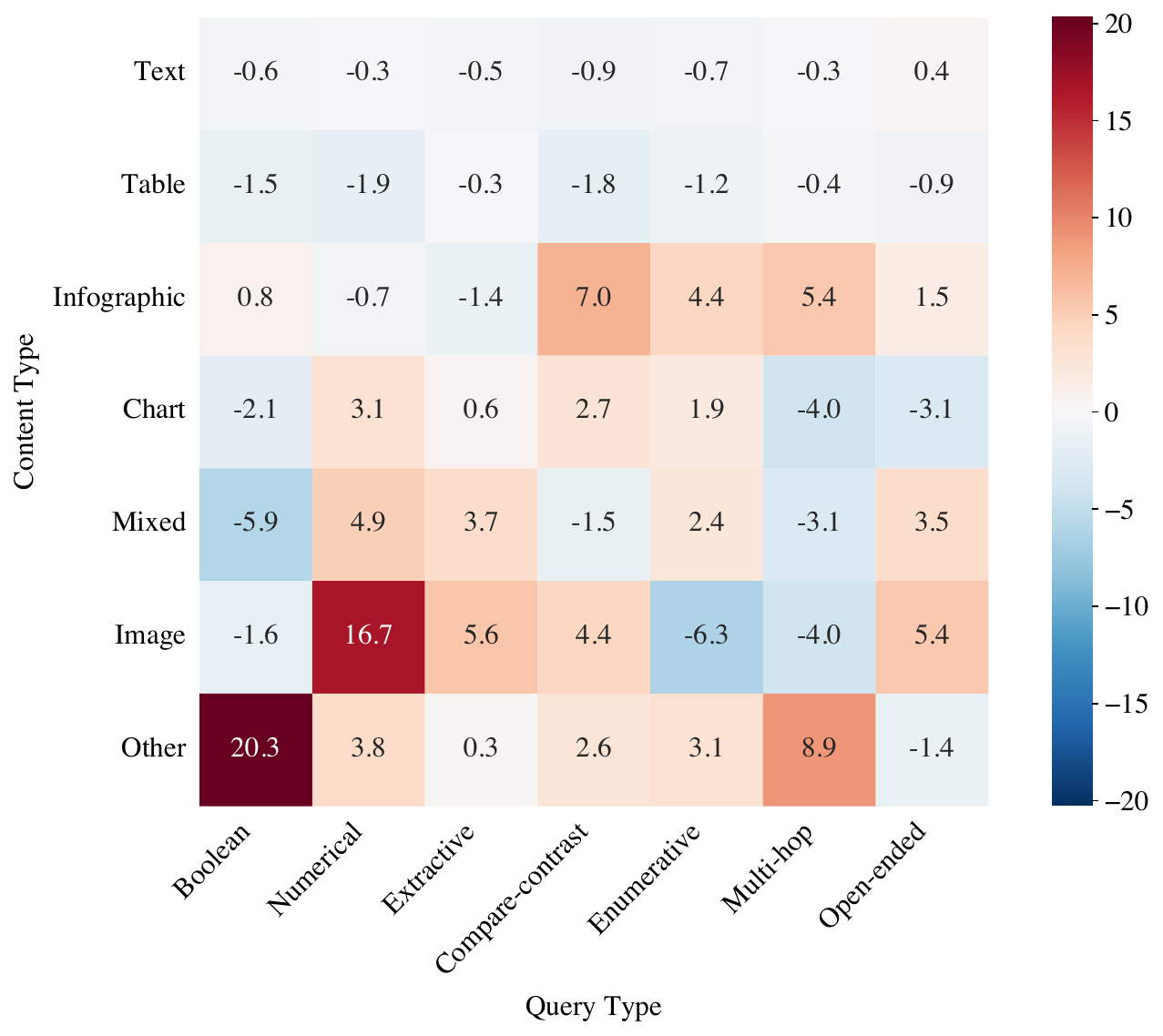}
\caption{\textbf{Residuals from additive performance model.} Each cell shows the difference between observed NDCG@10 and the value predicted by an additive model of query type and content type main effects. Values near zero (white) indicate no interaction; positive values (red) indicate better-than-expected performance for that combination; negative values (blue) indicate worse-than-expected.} 
\label{fig:ndcg content query residuals heatmap}
\end{figure}

NDCG@10 by content type in Table~\ref{tab:perf content type} show that retrieval is more challenging for visual content, with Image performing 10pp below Text. However, content type and query type are correlated in our benchmark: for instance, tables appear in numerical queries 2.2 $\times$ more often than the baseline, while images are over-represented in open-ended queries (Figure~\ref{fig:lift content type}). Since numerical queries are easier than open-ended ones, we test whether the effect of content type is a byproduct of query type confounding. We fit an additive model that predicts performance as the sum of independent query-type and content-type effects. Figure~\ref{fig:ndcg content query residuals heatmap} shows the residuals which measure deviation from this baseline. We see that most residuals are below 5pp, indicating that the two factors combine additively without significant interaction.

\begin{table}[h!]
\small
\centering
\begin{tabular}{lcc}
\toprule
\textbf{Content type} & NDCG@10 & Content type count \\
\midrule
Text & 59.3 & 17244 \\
Chart & 56.3 & 2364 \\
Infographic & 55.2 & 2814 \\
Table & 53.9 & 6480 \\
Other & 50.8 & 492 \\
Image & 49.3 & 1140 \\
Mixed & 45.1 & 1164 \\
\bottomrule
\end{tabular}
\caption{\textbf{ColEmbed-3B-v2 NDCG@10 by content type.} Content type is labeled on each annotated page based on the nature of the query-relevant content delimited by the bounding boxes. One page may be tagged with several content types if it contains multiple relevant sections of distinct nature. The Mixed type corresponds to annotations encompassing several content types.}
\label{tab:perf content type}
\end{table}

\section{VLM Filtering Effect on Query Distribution} 
\label{appendix: vlm filtering stats}

Table~\ref{tab:vlm_filtering_model_selection} reports the recall/precision trade-off that motivated the choice of Qwen2.5-VL-32B-Instruct as the pre-filtering model.

\begin{table}[h!]
\centering
\small
\begin{tabular}{lcc}
\toprule
\textbf{Model} & \textbf{Recall} & \textbf{Precision} \\
\midrule
Qwen2.5-VL-32B & 98\% & 32\% \\
Gemini-2.5-Flash & 78\% & 81\% \\
\bottomrule
\end{tabular}
\caption{\textbf{VLM pre-filtering model selection.} Evaluated on 697 human-annotated query--page pairs from ViDoRe v2~\citep{macé2025vidorebenchmarkv2raising} (103 positives, 594 negatives). We selected Qwen2.5-VL-32B as a high-recall sieve to avoid discarding $\sim$25\% of relevant documents, leaving final relevance judgments to human annotators.}
\label{tab:vlm_filtering_model_selection}
\end{table}

To assess whether VLM pre-filtering introduces distributional bias, we compare query type, query format, and page modality distributions before and after the filtering step.

Table~\ref{tab:filtering_query_type} shows that filtering causes minimal shifts across query types (all $|\Delta| \leq 3\%$), with the exception of Open-Ended queries ($-5.4\%$), which reflects the correct removal of overly vague queries. Table~\ref{tab:filtering_query_format} shows similarly small shifts across query formats. Table~\ref{tab:filtering_modality} shows that the modality distribution of relevant pages is close to the source documents, confirming that complex visual content is not systematically excluded. The slight decrease in graphical elements use occurs because semantically irrelevant decorative items, such as logos, are naturally not targeted by queries.

\begin{table}[h!]
\centering
\small
\begin{tabular}{lccc}
\toprule
\textbf{Query Type} & \textbf{Before (\%)} & \textbf{After (\%)} & ${\Delta}$ \\
\midrule
Boolean           & 13.88 & 16.07 & $+2.20$ \\
Compare-Contrast  & 13.73 & 13.25 & $-0.47$ \\
Enumerative       & 10.31 & 10.16 & $-0.15$ \\
Extractive        & 17.17 & 20.11 & $+2.94$ \\
Multi-Hop         &  8.27 &  8.06 & $-0.21$ \\
Numerical         &  6.32 &  7.42 & $+1.10$ \\
Open-Ended        & 30.33 & 24.93 & $-5.41$ \\
\bottomrule
\end{tabular}
\caption{\textbf{Query type distribution before and after VLM filtering.}}
\label{tab:filtering_query_type}
\end{table}

\begin{table}[h!]
\centering
\small
\begin{tabular}{lccc}
\toprule
\textbf{Query Format} & \textbf{Before (\%)} & \textbf{After (\%)} & ${\Delta}$ \\
\midrule
Instruction & 31.32 & 30.50 & $-0.83$ \\
Keyword     & 20.00 & 17.51 & $-2.49$ \\
Question    & 48.67 & 51.99 & $+3.32$ \\
\bottomrule
\end{tabular}
\caption{\textbf{Query format distribution before and after VLM filtering.}}
\label{tab:filtering_query_format}
\end{table}

\begin{table}[h!]
\centering
\small
\begin{tabular}{lccc}
\toprule
\textbf{Source} & \textbf{Text} & \textbf{Table} & \textbf{Graphical} \\
\midrule
Relevant Pages  & 55.52\% & 21.20\% & 23.28\% \\
Original Docs   & 54.10\% & 15.81\% & 30.09\% \\
\bottomrule
\end{tabular}
\caption{\textbf{Modality distribution for relevant pages vs.\ original documents.} The proportion of text, table, and graphical elements is preserved after filtering, confirming that complex visual content is not systematically excluded.}
\label{tab:filtering_modality}
\end{table}

\section{Bounding box annotations}
\label{appendix: bbox}

\paragraph{Inter-annotator agreement} \label{appendix: annotator agreement bbox}

Table~\ref{tab:inter-annotator bbox agreement} shows IoU and F1 scores between human annotations, to detail results of Section~\ref{sec: grounded qa}.

\begin{table*}[h!]
\centering
\resizebox{\textwidth}{!}{%
\begin{tabular}{l c c c c c c c c c c >{\columncolor{avgcol}}c}
\toprule
 & \multicolumn{7}{c}{\textbf{English Datasets}} & \multicolumn{3}{c}{\textbf{French Datasets}} & \multicolumn{1}{c}{} \\
\cmidrule(lr){2-8} \cmidrule(lr){9-11}
\textbf{Metric} & C.S. & Nucl. & Fin. & Phar. & H.R. & Ind. & Tele. & Phys. & Ener. & Fin. & \textbf{Average} \\
\midrule
\textbf{IoU} & 0.500 & 0.476 & 0.462 & 0.615 & 0.474 & 0.502 & 0.526 & 0.443 & 0.470 & 0.503 & \textbf{0.497} \\
\textbf{F1} & 0.608 & 0.594 & 0.569 & 0.720 & 0.594 & 0.611 & 0.637 & 0.540 & 0.569 & 0.581 & \textbf{0.602} \\
\bottomrule
\end{tabular}
}
\caption{\textbf{Inter-annotator agreement metrics on bounding box annotations.}}
\label{tab:inter-annotator bbox agreement} 
\end{table*}

\paragraph{Validation on high-agreement pages}
To further characterize model grounding performance, we restrict evaluation to query--page pairs with high inter-annotator agreement (mean pairwise IoU $\geq 0.7$). On this subset, Qwen3-VL-30B-A3B-Instruct and Gemini 3 Pro achieve both substantially higher than full-set scores, yet still well below human performance (\ref{tab:grounding_results}). This confirms that visual grounding is a genuine open challenge in current models.

\begin{table}[h!]
    \small
    \centering
    \resizebox{\columnwidth}{!}{
    \begin{tabular}{@{}llcc@{}}
        \toprule
        \textbf{Model} & \textbf{Evaluation Set} & \textbf{F1} & \textbf{IoU} \\ \midrule
        Qwen3-VL-30B-A3B-Instruct & Full Set & 0.089 & — \\
        & High-Agreement & \textbf{0.311} & \textbf{0.262} \\ \addlinespace
        Gemini 3 Pro & Full Set & 0.065 & — \\
        & High-Agreement & \textbf{0.208} & \textbf{0.155} \\ \bottomrule
    \end{tabular}
    }
    \caption{Model grounding performance comparison between the full evaluation set and the high-agreement subset (mean pairwise IoU $\geq 0.7$).}
    \label{tab:grounding_results}
\end{table}

\paragraph{Bounding box predictions}

\begin{figure*}[h!]
    \centering
    \includegraphics[width=\textwidth]{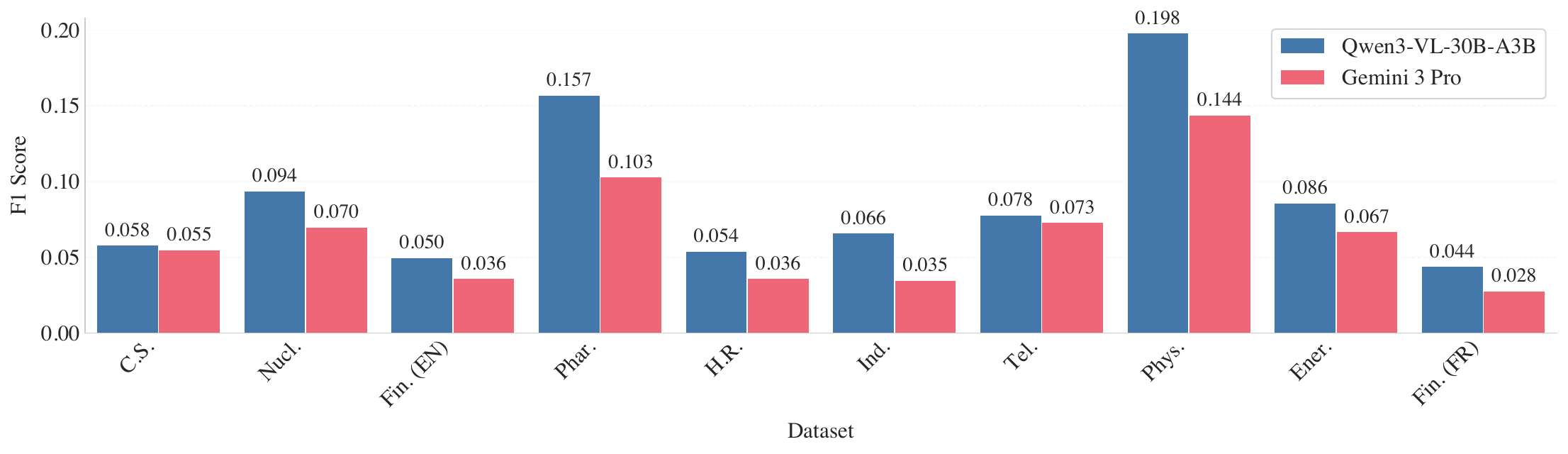}
    \caption{\textbf{Model bounding box localization performance}. Each F1 score measures the zone-based overlap between model-generated bounding boxes and human annotations, using the annotator yielding the highest F1.} 
    \label{fig:bbox evaluation}
\end{figure*}

Figure~\ref{fig:bbox_instructions} shows the prompt used to generate final answers with inline bounding boxes for visual grounding, and Figure~\ref{fig:bbox evaluation} reports bounding box localization F1 scores by dataset.

\section{Final answer evaluation} \label{appendix:final answer eval}

\paragraph{Evaluation setup} Generated final answers are evaluated in a pass@1 setting using GPT 5.2 with medium reasoning effort as the LLM judge. The judge compares each generated answer against the ground-truth annotation and returns a binary correctness label. The answer generation and judge prompts are shown in Figure~\ref{fig:generation_prompt} and Figure~\ref{fig:judge_prompt} respectively. We evaluated Gemini 3 Pro with low thinking effort, GPT-5 with medium reasoning effort, as well as the thinking version of Qwen3-VL-235B-A22B.

To assess the reliability of our judge, we conducted 5 independent evaluation runs on a fixed set of Gemini 3 Pro outputs. Individual run scores showed minimal fluctuation (mean 72.09\,\%, $\sigma=0.22$\,\%) and high internal consistency (Krippendorff's $\alpha = 0.91$), confirming that the judge is consistent given a fixed context.

\paragraph{End-to-End Pipeline Stability} While the judge demonstrates high consistency on fixed inputs, the full evaluation pipeline introduces a second layer of variability: the model's generation process. To quantify the end-to-end variance under rigorous conditions, we performed 5 independent runs. For computational efficiency, we restricted this stress test to the most challenging corpus in each language: \textit{Industrial Maintenance} (English) and \textit{Finance} (French).

We measured an average score of 65.74\,\% with a standard deviation of 0.94\,\%. Crucially, the evaluation signal remains robust against generative noise, achieving a Krippendorff's $\alpha$ of 0.80. This agreement confirms that the end-to-end results are statistically reliable even when subjected to the most difficult evaluation scenarios.

\paragraph{Easy/hard query filtering} To classify queries by difficulty, we prompt a panel of 6 LLMs to answer each query without access to any corpus context. We select GPT-5-nano, GPT-5-mini, GPT-5, Qwen3-VL-30B-A3B, Gemini 2.5 Flash, and Gemini 2.5 Pro to span different model families and capability levels. Each model receives only the query text and is asked to provide a direct answer with the prompt in Figure~\ref{fig:easy_query_prompt}. Answers are evaluated for correctness using the same GPT-5.2 judge described above. A query is labeled \textit{easy} if at least one model answers correctly, and \textit{hard} otherwise. Table~\ref{tab: easy hard filtering} reports per-model accuracy and the resulting proportion of easy queries for each dataset. The distribution varies substantially across domains: knowledge-intensive datasets such as Computer Science and Physics have over 85\% easy queries, while domain-specific datasets such as Finance and Energy contain fewer than 35\% easy queries, reflecting the specialized nature of their content.

\begin{table*}[h!]
\centering
\label{tab:performance_model}
\begin{tabular}{l c c c c c c c c >{\columncolor{avgcol}}c}
\toprule
 & \multicolumn{5}{c}{\textbf{English Datasets}} & \multicolumn{3}{c}{\textbf{French Datasets}} & \multicolumn{1}{c}{} \\
\cmidrule(lr){2-6} \cmidrule(lr){7-9}
\textbf{Model} & C.S. & Fin. & Phar. & H.R. & Ind. & Phys. & Ener. & Fin. & Total \\
\midrule
GPT-5-nano     & 74.4 & 7.4  & 30.5 & 12.9 & 15.6 & 74.2 & 14.0 & 9.1  & 29.8 \\
GPT-5-mini      & 79.1 & 13.3 & 37.4 & 17.6 & 20.5 & 80.1 & 13.6 & 13.1 & 34.3 \\
GPT-5           & 76.3 & 25.2 & 50.8 & 29.9 & 32.2 & 80.5 & 26.0 & 22.2 & 42.9 \\
Qwen3-VL-30B-A3B & 60.9 & 3.9  & 19.2 & 6.3  & 9.9  & 60.9 & 6.8  & 4.4  & 21.5 \\
Gemini 2.5 Flash& 66.1 & 8.7  & 30.8 & 13.5 & 15.9 & 63.6 & 14.9 & 13.1 & 28.3 \\
Gemini 2.5 Pro  & 70.2 & 16.8 & 29.4 & 15.4 & 23.7 & 63.9 & 20.5 & 20.0 & 32.9 \\
\midrule
\textbf{Easy queries (\%)}           & 86.5 & 31.7 & 57.1 & 36.5 & 38.9 & 86.4 & 32.5 & 30.0 & 48.6 \\
\bottomrule
\end{tabular}
\caption{\textbf{Percentage of queries correctly answered by LLMs without corpus context.} A panel of 6 LLMs is asked to answer the queries of the 8 public datasets without access to any corpus context. Queries correctly answered by at least one of the 6 models are classified as \textit{easy} queries, while the rest are labeled as \textit{hard}. Easy queries account for 48.6\,\% of all the queries.}
\label{tab: easy hard filtering}
\end{table*}

\twocolumn

\clearpage

\section{Visual grounding examples}

\begin{figure*}[h!]
    \centering
    \includegraphics[width=\textwidth]{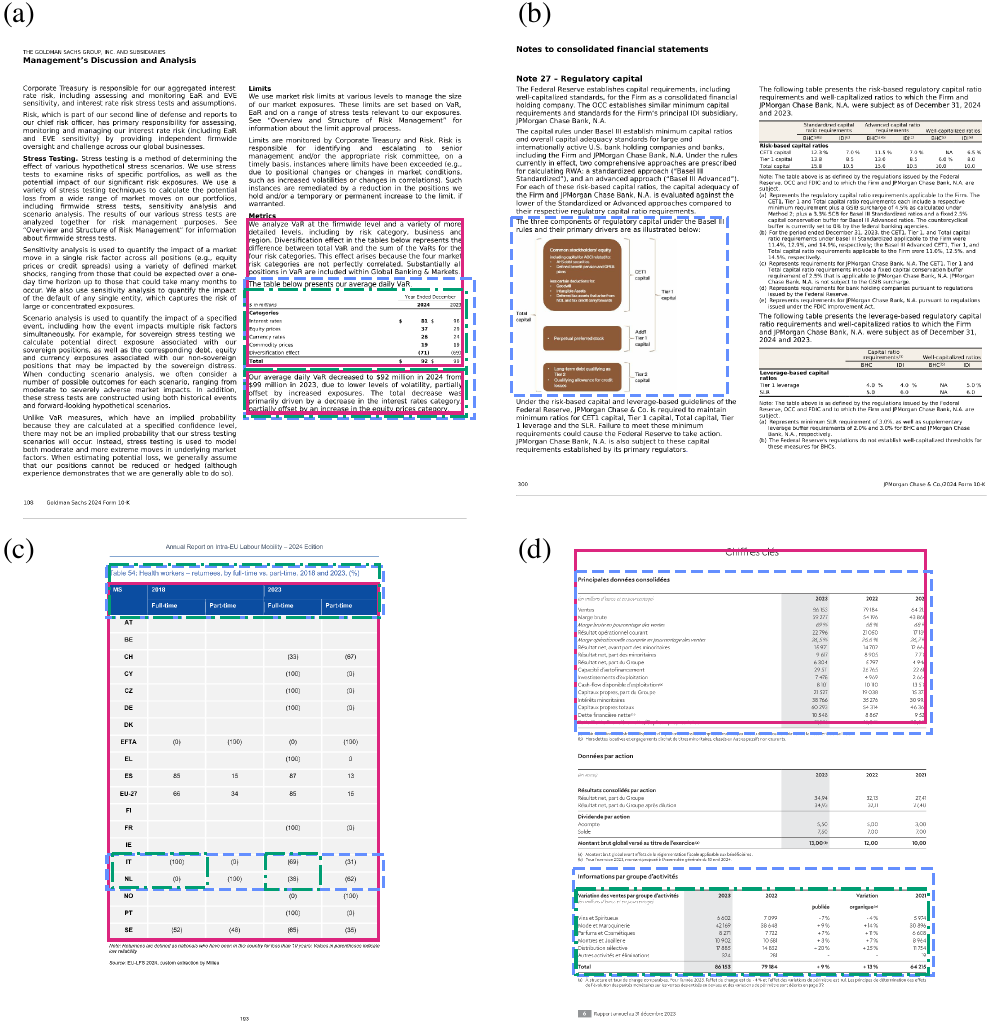}
    \caption{\textbf{Visual grounding comparative examples for Qwen3-VL-30B-A3B.} Each panel shows a document page with Qwen3-VL's predicted bounding boxes (solid magenta) and human bounding boxes (dashed blue and green, one color per annotator). Corresponding datasets and queries: (a) finance\_en: What was the average daily Value at Risk (VaR) for Goldman Sachs during 2024?, (b) finance\_en: List the 3 components of regulatory capital under Basel III, and determine the role of each component., (c) hr\_en: Analyze how full-time employment among returning health workers evolved in the Netherlands and Italy from 2018 to 2023, and describe the differences in their employment trends., (d) finance\_fr: Croissance Mode Maroquinerie vs Vins Spiritueux 2023 performance}
    \label{fig:bbox examples qwen}
\end{figure*}

\begin{figure*}[h!]
    \centering
    \includegraphics[width=\textwidth]{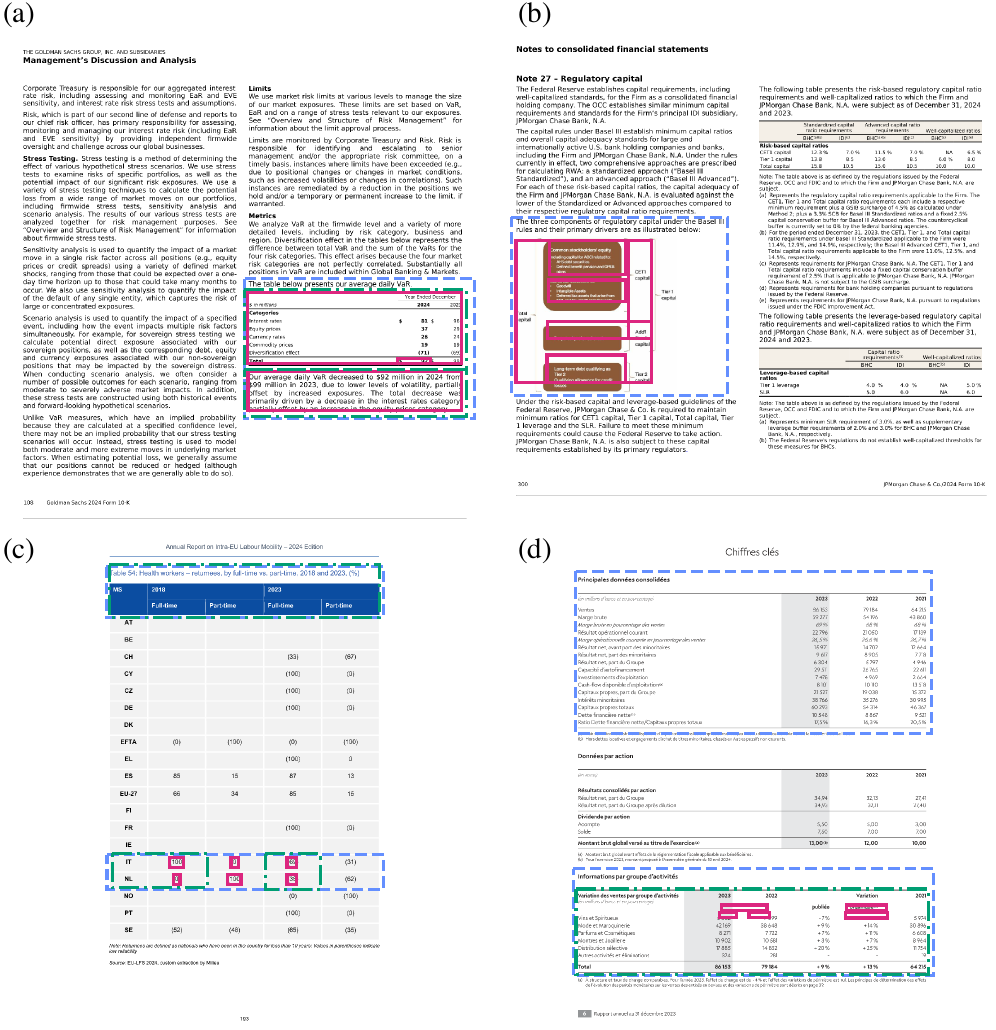}
    \caption{\textbf{Visual grounding comparative examples for Gemini 3 Pro.} Each panel shows a document page with Gemini's predicted bounding boxes (solid magenta) and human bounding boxes (dashed blue and green, one color per annotator). Corresponding datasets and queries: (a) finance\_en: What was the average daily Value at Risk (VaR) for Goldman Sachs during 2024?, (b) finance\_en: List the 3 components of regulatory capital under Basel III, and determine the role of each component., (c) hr\_en: Analyze how full-time employment among returning health workers evolved in the Netherlands and Italy from 2018 to 2023, and describe the differences in their employment trends., (d) finance\_fr: Croissance Mode Maroquinerie vs Vins Spiritueux 2023 performance}
    \label{fig:bbox examples gemini}
\end{figure*}

Qualitative analysis reveals distinct failure modes. Gemini frequently produces off-by-one page indexing errors: the predicted coordinates would correctly localize the target content if applied to an adjacent page. The two models also differ in box granularity: Gemini tends to draw tight boxes around individual elements (e.g., a single table cell or text line), whereas Qwen3-VL generates larger boxes encompassing entire sections or paragraphs, more closely matching human annotation patterns. Figures~\ref{fig:bbox examples qwen} and~\ref{fig:bbox examples gemini} illustrate these tendencies across four dataset pages: Qwen3-VL's bounding boxes are comparatively wide and encompass entire page elements (pages (a), (c), and (d)), while Gemini 3 Pro's visual grounding is more precise (pages (b) and (c)). This difference in granularity partially explains Qwen3-VL's higher F1 scores, as broader boxes are more likely to overlap with the ground-truth zones used in our evaluation. Both models exhibit errors and omissions: in page (b), the chart is not labeled by Qwen3-VL, and in page (d), Gemini 3 Pro predicts incorrect bounding boxes for the bottom table while Qwen3-VL provides grounding for the wrong table.

\label{appendix: bbox examples}

\section{Instructions given to Annotators}
\label{appendix: annotator_instructions}

\paragraph{Query Generation}\Cref{fig:task_1_instructions} details step-by-step instructions to annotators to generate queries from summaries and images.

\begin{promptfigure}[blue!5]{\textbf{Instructions given to human annotators to create queries}}{fig:task_1_instructions}
\begin{verbatim}
# Step 1
    The annotator will be provided with content(text summary or image(s)) and a list of instructions
    on the queries that are expected.

# Step 2
    Read and analyse the content.

# Step 3
    - The annotator writes a series of queries that can, supposedly, be answered by the summarized
    content. It is okay if the information needed to answer the question is in other parts of the
    document or not explicitly written in the summary.
    - If the summary is not adapted to a specific type or format of questions, the annotator may skip. 
    - They should follow the number of queries asked for each category.
    - They should follow the expected type of queries provided and the format.
\end{verbatim}
\end{promptfigure}

\paragraph{Query-Page Relevancy linking}\Cref{fig:task_2_instructions} details the step-by-step instructions provided to annotators for assessing page relevance, identifying content modalities, and localizing evidence via bounding boxes. \Cref{tab:relevance_scale} gives the definitions of relevancy scores used by the human annotators.

\begin{table}[h!]
\centering
\small
\begin{tabularx}{\columnwidth}{clX} 
\toprule
\textbf{Score} & \textbf{Label} & \textbf{Definition} \\
\midrule
2 & Fully Relevant & The page contains the complete answer. \\ 
\addlinespace % Adds a nice breathing room between rows without breaking lines
1 & Critically Relevant & The page contains facts or information required to answer the query, though additional information is required. \\ 
\addlinespace
0 & Not Relevant & Provides no information relevant to the query. \\
\bottomrule
\end{tabularx}
\caption{\textbf{Relevance definitions used for page-level annotations.}}
\label{tab:relevance_scale}
\end{table}

\begin{promptfigure}[blue!5]{\textbf{Instructions given to human annotators to annotate query-page relevancy}}{fig:task_2_instructions}
\begin{verbatim}
# Task overview
In this task, the annotator will be provided with a query and pages that are supposed to be relevant
to answer the query. The annotator’s goal is to rate the relevance in answerability of each page with
respect to the query.

# Step 1:
    - Review the query and pages to get an understanding of the content and domain
    
# Step 2:
    Rate the query quality. 
    - If adheres to guidelines > Good (1)
    - If doesn’t adhere to guidelines> Poor(0)
    - If the query is Poor quality, skip the task.
    
# Step 3:
    - For each page, rate the relevance with respect to the query
    - If page completely answer query > Fully Relevant(2)
    - If page contains information required to answer the query > Critically Relevant(1)
    - If page contains no relevant information > Not Relevant(0)
    
# Step 4:
    - For each page, annotate the modalities in which the relevant information is located concerning
    the query-page link, relevant information can be located in multiple modalities at the same time:
Modality : [“text”, “table”, “chart”, infographic”, “image”, “other”]
    - If relevance score = 0, modality may be “N/A”
    
# Step 5:
    - For each page, draw bounding boxes around the relevant text/chart/image/infographic (if any)
    - If relevance score = 0, do not draw a bounding box
    
# Step 6:
    - Repeat steps 3, 4 and 5 for all pages
    
# Step 7:
    - Propose an answer to the query, given the relevant pages.
    - If the query is not answerable, rate it as “unanswerable”

\end{verbatim}
\end{promptfigure}

\section{Prompts}

All the prompts used for both dataset generation and evaluations are detailed from \Cref{fig:query_generation_prompt} to \Cref{fig:bbox_instructions}.

\begin{promptfigure}{\textbf{Prompt used generate synthetic queries with the NeMo Data Designer tool}}{fig:query_generation_prompt}
\begin{verbatim}
<mission>
You are an assistant specialized in visual document understanding tasks. You will be given a context, 
summarizing the content of a section or multiple document sections. Your goal is to carefully analyze 
the context and to solve a series of tasks related to its content. You are tasked with generating 
query-answer pairs. Your queries will be used to simulate a user unfamiliar with the specific content 
of the page, and who is looking for information in a large knowledge base through a search engine. 
The user does not have access to the document and is looking for information that can be present in 
any document in the knowledge base.
</mission>

<definitions>
- A query is said to be fully answerable if the page contains a precise and complete answer to the 
query.
- A query is said to be partially answerable if the page contains relevant information that is 
directly related to the query but some key information is missing and must be retrieved in other 
pages or documents in order to give a precise and complete answer.
- An open-ended query is an explanatory or descriptive query that synthesizes  information; may be 
broad in scope and focused on qualitative aspects of the summary
- A compare-contrast query is a query that requires comparing and/or contrasting multiple entities or 
topics that are closely related to each other
- An enumerative query is a query that asks to list all examples that possess a common property, 
optionally requesting details about the specifics of each example.
- A numerical query is a query that asks for a specific number or calculated number given a summary. 
The query should require more than simply reading numbers directly from the page.
- A boolean query is a yes/no query that may involve multiple steps of reasoning.
- An extractive query is a clear and specific query that can be answered using only a specific piece 
of information.
- A multi-hop query is a complex query that requires retrieving and integrating information from 
multiple sources or steps to produce a complete answer.
- A question query is a complete sentence that ends with a question mark, typically used to seek 
specific information or clarification.
- A keyword query is a brief, often fragmented phrase or set of terms used to search or filter 
information, without forming a full grammatical sentence.
- An instruction query is a directive that describes a task to be performed on the documents, often in 
the form of a command or request.
</definitions>
<rules>
<queries>
- Generate queries only in {{ language }}.
- Make queries diverse, natural, and plausible for someone unfamiliar with the document.
- Each query must be standalone; do not reference “the page”, “the table”, “the figure”, 
“the document”, “the text”, “the table of contents”, etc.
- Rephrase; avoid copying wording from the source so semantic matching, not surface matching, 
is tested.
- You may include queries about relationships or trends often shown in tables/figures/graphs, 
but never refer to a specific table/figure.
- Avoid overly generic queries that apply to any document.
- Keep each query concise ({{ length }} words).
- When appropriate, write multi-hop queries that integrate information across the provided pages.
</queries>
</rules>
<instructions>

Used Documents: {{ document_names }}
<summary>
{{summary}}
</summary>
Using the provided context, generate a {{ difficulty }}, {{ reasoning_type }}, {{ answerability }}
query. The query should be {{ query_type }} using the provided context and have the format of a 
{{ query_format }} query.
The query should be self-sufficient and related to the context.
</instructions>
\end{verbatim}
\end{promptfigure}

\begin{promptfigure}{\textbf{Prompt used to pre-filter the irrelevant pages for a given query}}{fig:query_linking_prompt}

\begin{verbatim}
<mission>
You are an assistant specialized in visual document understanding tasks. You will be given a document
page by page and a question. Your goal is to carefully analyze the page and say if it is related to
the question's answer. You are tasked with generating question-page affiliation as well as the
question answer if it exists in the page. 
</mission>

<definitions>
- A question is said to be fully answerable if the corresponding page contains a precise and complete
answer to the question.
- A question is said to be partially answerable if the corresponding page content is necessary to
answer the question but some key information is missing.
- A question is said to be unanswerable if the corresponding page contains information related to the
question's topic or domain but upon closer inspection does not contain information that is useful to
answer the question. Or if the page has no link whatsoever with the query.
</definitions>

<rules>
<page_affiliation>
- Be sure to put the relevance (and only that) between the tags <relevance>...</relevance>.
The possible values are: 
    <relevance>fully answerable</relevance>,
    <relevance>partially answerable</relevance>
    <relevance>unanswerable</relevance>.
- Be very careful when doing your page affiliation. Only say a page is relevant when it really is.
</page_affiliation>
<answers>
- You must generate the answer between the tags <answer>...</answer>. Between these tags, you should
only put the answer to the question.
- You must generate answers in the following language: {language}
- Your answers should be complete sentences.
- When the question is ambiguous, your answer should state that there is an ambiguity in the question.
- You should always generate the answer based on all the information available on the page, even if
the question was generated only on part of the page.
</answers>
</rules>

<instructions>
Return if the following question is "fully answerable" "partially answerable" or "unanswerable" based
on the content of the page between the tags <relevance> ... </relevance>. 
If the question is answerable, provide the answer.

Here is the question :
{{ query }}

And there is the page content :

</instructions>
\end{verbatim}
\end{promptfigure}

\begin{promptfigure}{\textbf{Prompt used to merge human annotators answers}}{fig:answer_merging_prompt}
\begin{verbatim}
You are given a set of document pages (images), a query, and a list of one or more
proposed answers.

Query :

{{ query }}

Proposed Answers:

{{ answers }}

Your task is to carefully analyze the provided pages, the query, and the proposed answers. You must
return a single, syntactically correct JSON object with the following structure:
```json
{
    "reasoning": "<string>",
    "information_in_pages": <true or false>,
    "answer_correctness": [<true or false>, ...],
    "reformulated_answer": "<string>"
}
```
Instructions for each field:
- reasoning: Explain the logic for each boolean in the `answer_correctness` list.
For each proposed answer, state why it is correct or incorrect, citing specific  evidence from the
document pages.

- information_in_pages: Set to `true` if the information needed to definitively answer the query
is present in the pages. Otherwise, set to `false`.

- answer_correctness: A list of booleans, corresponding to each proposed answer  in the original
order. Use `true` if the answer is verifiably correct based on the pages and `false` otherwise.

- reformulated_answer: A single string containing the most precise and correct answer to the query,
derived only from information in the pages. If any of the proposed answers are correct, use them as
a basis for synthesizing this improved answer. The reformulation must be concise and factual.

Important rules:
- Base your entire analysis strictly on the content of the provided document pages. 
Do not use outside knowledge.

- Do not invent, infer, or assume information that is not explicitly stated in the pages.

- Always provide a string for the `reformulated_answer`, even if no correct answer can be formed
from the text.

- Your final output must be only the JSON object.
\end{verbatim}
\end{promptfigure}

\begin{promptfigure}{\textbf{Easy/hard query filtering prompt}}{fig:easy_query_prompt}
\begin{verbatim}
Give a very precise and concise answer to the following query.
If you are unable to answer, output 'I don't know'.
Query: {{ query }}
\end{verbatim}
\end{promptfigure}

\begin{promptfigure}{\textbf{Judge prompt used for end to end evaluation}}{fig:judge_prompt}
\begin{verbatim}
You are an expert judge evaluating the accuracy of a test answer against a gold-standard true answer.
Your goal is to determine if the test answer captures the essential "core information."

### Evaluation Criteria:
- Correct: The test answer contains all core information of the true answer. Minor omissions of
non-essential details or the addition of minor, non-contradictory information should still be marked
as "Correct."
- Partially Correct: The test answer captures some of the core information, but suffers from
significant omissions or includes substantial extra information that was not requested or present in
the true answer.
- Incorrect: The test answer is fundamentally wrong, contradicts the true answer, or misses the core
information entirely.

### Input Data:
Query: {{ query }}

True Answer: {{ true_answer} }

Test Answer: {{ test_answer }}

### Output Format:
Provide a very brief explanation for your judgment. You must output your final response in a JSON
format with two fields: "explanation" and "judgment" (which must be "Correct", "Partially Correct",
or "Incorrect").
\end{verbatim}
\end{promptfigure}

\begin{promptfigure}{\textbf{Answer generation prompt used for end to end evaluation}}{fig:generation_prompt}
\begin{verbatim}
You are an expert at answering query based on documents.

Here is a list of relevant documents: {{ documents }}

Based on the above documents, answer the following query: {{ query }}

Keep the response short when appropriate. Output the answer only.
\end{verbatim}
\end{promptfigure}

\begin{promptfigure}{\textbf{Query translation from English to French prompt}}{fig:query_translation}
\begin{verbatim}
English text to translate: {{ query }}

Translate the English text above to French.
Make sure you follow the format of the English text. Don't change acronyms.
Follow the following json schema.
{
    "french_translation": ...
}
\end{verbatim}
\end{promptfigure}

\begin{promptfigure}{\textbf{Bounding box prediction prompt}}{fig:bbox_instructions}
\begin{verbatim}
# Role and Objective
- Serve as an expert in document analysis and visual grounding.
- Given a query and multiple document page images, provide a natural language answer with inline
grounding references.

# Instructions
- Analyze all provided pages to answer the query comprehensively.
- For each piece of information used in your answer, provide visual grounding by including bounding
box coordinates of all the sections of the document that help answer the query.
- Use this format to include the list of all bounding boxes of image N:
<bboxes image="N">[[x_{min}, y_{min}, x_{max}, y_{max}], ...]</bboxes>
  - image="N" specifies the 0-indexed page number (0=first page, 1=second page, etc.)
  - Include bounding boxes inline in your answer, immediately after mentioning the relevant
  information
  - A given page may contain multiple non-contiguous sections that help answer the query. In this
  case, you must output the list of the bounding boxes of all these sections.
  - You must group all the bounding boxes of a given page into a single 
    <bboxes image="N">...</bboxes> tag.

# Grounding Principles
- DO NOT output more than 5 bounding boxes per page.
- Adjacent logical units must be enclosed in a single, continuous bounding box.
- Return multiple bounding boxes only if information is clearly independent and separated by
significant non-relevant content.

# Output Format
- Provide a natural language answer to the query.
- Embed grounding tags directly inline where relevant information is discussed.
- Example: "The valuation technique described on page 1 
  <bboxes image="0">[[120, 450, 890, 670], [100, 800, 330, 960]]</bboxes> 
  uses discounted cash flow analysis."
\end{verbatim}
\end{promptfigure}

\end{document}